\documentclass{article}

\usepackage{arxiv}

\usepackage[utf8]{inputenc} % allow utf-8 input
\usepackage[T1]{fontenc}    % use 8-bit T1 fonts
\usepackage{hyperref}       % hyperlinks
\usepackage{url}            % simple URL typesetting
\usepackage{booktabs}       % professional-quality tables
\usepackage{amsfonts}       % blackboard math symbols
\usepackage{nicefrac}       % compact symbols for 1/2, etc.
\usepackage{microtype}      % microtypography
\usepackage{lipsum}		% Can be removed after putting your text content
\usepackage{graphicx}
\usepackage{natbib}
\usepackage{doi}
\usepackage{amsmath}
\usepackage{adjustbox}
\usepackage{algorithm}
\usepackage{algorithmic}

\title{Exploring the Magnitude-Shape Plot Framework for Anomaly Detection in Crowded Video Scenes}

\author{ Zuzheng~Wang, Fouzi~Harrou, Ying~Sun, Marc~G~Genton \\
	King Abdullah University of Science and Technology (KAUST)\\
	Computer, Electrical and Mathematical Sciences and Engineering (CEMSE) Division\\
	Thuwal 23955-6900, Saudi Arabia \\
	\texttt{(Zuzheng.Wang, fouzi.harrou, ying.sun, marc.genton) @kaust.edu.sa} \\
	}

\renewcommand{\shorttitle}{\textit{arXiv} Template}

\hypersetup{
pdftitle={A template for the arxiv style},
pdfsubject={q-bio.NC, q-bio.QM},
pdfauthor={David S.~Hippocampus, Elias D.~Striatum},
pdfkeywords={First keyword, Second keyword, More},
}

\begin{document}
\maketitle

\begin{abstract}
	Detecting anomalies in crowded video scenes is critical for public safety, enabling timely identification of potential threats. This study explores video anomaly detection within a Functional Data Analysis framework, focusing on the application of the Magnitude-Shape (MS) Plot. Autoencoders are used to learn and reconstruct normal behavioral patterns from anomaly-free training data, resulting in low reconstruction errors for normal frames and higher errors for frames with potential anomalies. The reconstruction error matrix for each frame is treated as multivariate functional data, with the MS-Plot applied to analyze both magnitude and shape deviations, enhancing the accuracy of anomaly detection. Using its capacity to evaluate the magnitude and shape of deviations, the MS-Plot offers a statistically principled and interpretable framework for anomaly detection. The proposed methodology is evaluated on two widely used benchmark datasets, UCSD Ped2 and CUHK Avenue, demonstrating promising performance. It performs better than traditional univariate functional detectors (e.g., FBPlot, TVDMSS, Extremal Depth, and Outliergram) and several state-of-the-art methods. These results highlight the potential of the MS-Plot-based framework for effective anomaly detection in crowded video scenes.

\end{abstract}

% keywords can be removed

\section{Introduction} \label{sec1}

Anomaly detection in video sequences is a critical task in computer vision, particularly in scenarios involving crowded environments such as public spaces, events, transportation hubs, and marketplaces~\cite{santhosh2020anomaly,tripathi2018suspicious,chandola2009anomaly}. The goal is to identify unusual or potentially dangerous events that deviate from expected crowd behavior, such as sudden dispersals, confrontations, or individuals engaging in atypical activities~\cite{nayak2021comprehensive,mu2021abnormal}. Early detection of such anomalies is crucial for ensuring public safety, managing emergencies, and preventing incidents. In crowded areas, the task of anomaly detection becomes especially challenging due to the complexity and high dimensionality of spatiotemporal data. Video sequences in these environments consist of continuous frames with numerous individuals and objects interacting, creating intricate patterns of motion and behavior~\cite{duong2023deep}. This complexity requires the use of advanced methods capable of capturing not only spatial features within individual frames, but also temporal and contextual dependencies that characterize typical crowd dynamics~\cite{zhu2012context,pawar2019deep}.

\newpage
\medskip
Over the past two decades, researchers have developed numerous methods to effectively identify abnormal events in video data~\cite{morris2008survey}. Trajectory-based approaches track moving objects to detect deviations from typical patterns~\cite{kim2009observe,mehran2009abnormal,basharat2008learning}. For instance, the study in~\cite{piciarelli2008trajectory} explored anomaly detection through trajectory analysis using a single-class support vector machine (SVM) clustering approach, effectively identifying unusual trajectories without prior knowledge of outlier distributions. In~\cite{cocsar2016toward}, snapped trajectories are introduced as a high-level representation that reduces computational load and identifies key scene regions. This combined approach detects anomalies in speed, direction, and finer motion details with fewer false alarms. Trajectory-based methods rely on accurate detection and tracking but are limited by crowd density, resolution, motion, and occlusion, making them more suitable for sparse crowds. However, in crowded scenes, frequent occlusions, intersecting paths, and the challenge of distinguishing similar individuals~\cite{ma2014depth} reduce their effectiveness, and they lack contextual awareness for identifying anomalies.

\medskip
Alternatively, other techniques use spatiotemporal features to represent events in videos without requiring trajectory analysis. These methods capture motion dynamics by analyzing changes in pixel intensities and patterns over time~\cite{li2013anomaly}. Techniques such as spatiotemporal gradients~\cite{lu2013abnormal}, Histograms of Oriented Gradients (HOG)~\cite{dalal2005histograms}, 3D spatiotemporal gradient \cite{kratz2009anomaly}, and Histograms of Optical Flow (HOF)~\cite{dalal2006human} help detect collective crowd behaviors and abnormalities at the pixel or region level, enhancing anomaly detection in dense and dynamic scenes. For example, in~\cite{roy2018suspicious}, an SVM classifier trained on HOG features is proposed to automatically detect violent activities in surveillance videos, distinguishing actions like kicking and punching. This system triggers alerts for detected violence and monitors loitering duration to flag suspicious behavior in real-time, thus enhancing traditional surveillance effectiveness. In~\cite{ahad2018activity}, a method is introduced for recognizing overlapping and multi-dimensional actions using a spatiotemporal representation and enhanced Motion History Image (MHI). It employs Speeded-Up Robust Features (SURF) to capture key motion features, gradient-based optical flow for motion representation, and RANSAC (Random Sample Consensus) for outlier removal. The nearest neighbor classifier with leave-one-out cross-validation achieves improved recognition rates for complex actions in outdoor scenes compared to traditional MHI approaches. The study in~\cite{cheng2015video} introduces a hierarchical framework for video anomaly detection that integrates multi-level features, including 3D-SIFT (scale-invariant feature transform), HOF, and HOG, with Gaussian process regression. However, these low-level visual features (e.g., motion or texture) may still struggle to fully capture higher-level semantic information or complex contextual interactions in crowded scenes, limiting their effectiveness in more challenging anomaly detection scenarios~\cite{duong2023deep}. While effective in some cases, these methods often face limitations in generalizing across diverse environments and in capturing complex and subtle deviations from normal patterns.

\medskip
To address these limitations, deep learning-based methods have emerged as advanced alternatives for video anomaly detection, leveraging their capacity to capture high-level semantic information and complex interactions in crowded and dynamic environments~\cite{duong2023deep}. Unlike traditional techniques, deep learning models such as Convolutional Neural Networks (CNNs)~\cite{sabokrou2018deep,mansour2021intelligent,sultani2018real}, Recurrent Neural Networks (RNNs), and Long Short-Term Memory networks (LSTMs) can automatically learn and extract relevant spatial and temporal features directly from the data~\cite{nawaratne2019spatiotemporal,ullah2021efficient}. These models are particularly effective in modeling nuanced deviations from normal patterns, enabling them to generalize better across diverse scenarios. Moreover, autoencoders and Generative Adversarial Networks (GANs) are commonly used to reconstruct normal patterns in video frames, with deviations in reconstruction used as indicators of potential anomalies~\cite{luo2021future,micorek2024mulde,chen2022supervised}. The ability of deep learning models to handle complex feature representations has significantly improved the performance and adaptability of video anomaly detection systems across various application settings~\cite{nayak2021comprehensive,duong2023deep,pawar2019deep,wu2024deep}. The study in~\cite{sabokrou2018deep} presents a method for anomaly detection in crowded scenes using fully convolutional networks (FCNs) with temporal data. The model enhances feature extraction by combining a pre-trained CNN (a modified AlexNet) with a custom convolutional layer tailored to specific video data. In~\cite{zhou2016spatial}, a spatial-temporal CNN model is proposed for anomaly detection and localization using spatial-temporal volumes with motion information from static camera scenes. By capturing both appearance and motion features through spatial-temporal convolutions, the model enhances robustness. Evaluated on four benchmark datasets, it outperforms state-of-the-art methods, especially on challenging pixel-level criteria. In~\cite{hong2024making}, a video anomaly detection approach leverages a frame-to-label and motion (F2LM) generator to intentionally reduce the quality of abnormal regions, followed by a Destroyer that transforms these areas into zero vectors, making anomalies more prominent. This technique surpasses state-of-the-art performance on the UCSD Ped2, CUHK Avenue, and Shanghai Tech datasets. The study \cite{dilek2024enhancement} proposes an efficient frame-level video anomaly detection (VAD) method that utilizes transfer learning and fine-tuning on 20 CNN-based deep learning models, including variants of VGG, Xception, MobileNet, and ResNet.

\medskip 
Over the past decade, numerous reconstruction-based methods for video anomaly detection have emerged, leveraging models that learn to reconstruct normal video frames or sequences and using reconstruction errors to identify anomalies~\cite{chong2017abnormal,gong2019memorizing,le2023attention}.  The core idea of these approaches is that the model trained exclusively on anomaly-free data will struggle to accurately reconstruct anomalous events, resulting in higher reconstruction errors for frames with unusual or abnormal content. This discrepancy between normal and abnormal reconstructions provides a basis for detecting anomalies effectively. In~\cite{zhao2017spatio}, Zhao et al. tackle the challenge of detecting anomalies in complex video scenes by introducing a spatiotemporal AutoEncoder (STAE) that uses deep neural networks with 3D convolutions to learn spatial and temporal features. It incorporates a weight-decreasing prediction loss for future frame generation, improving motion feature learning beyond standard reconstruction loss. In \cite{deepak2021residual}, a residual spatiotemporal autoencoder (STAE) is proposed for video anomaly detection, where anomalies are identified as deviations from normal patterns using reconstruction loss. Residual connections enhance model performance, effectively reconstructing normal frames with low cost while detecting irregularities as abnormal frames.  Another study in \cite{luo2019video} proposed a sparse coding-inspired Deep Neural Network (DNN) for video anomaly detection, known as Temporally-coherent Sparse Coding (TSC). TSC maintains frame similarity using a temporal coherence term optimized via the Sequential Iterative Soft-Thresholding Algorithm (SIATA). The enhanced stacked Recurrent Neural Network Autoencoder (sRNN-AE) model introduces data-dependent similarity, reduced model depth for real-time detection, and temporal pooling for efficiency. The study in \cite{kumar2023efficient} considered a Residual Variational Autoencoder (RVAE) for unsupervised video anomaly detection, which captures complex patterns and minimizes reconstruction error through low-dimensional latent encoding and decoding. The model incorporates a ConvLSTM layer for improved spatiotemporal learning and uses residual connections to address the vanishing gradient problem. Recently, in \cite{aslam2024demaae}, a deep multiplicative attention-based autoencoder (DeMAAE) was introduced for video anomaly detection. DeMAAE applies a global attention mechanism at the decoder to enhance feature learning, leveraging an attention map created from encoder-decoder hidden states to guide decoding via a context vector. In \cite{le2023attention}, a spatial-temporal network achieved 97.4\% AUC on UCSD Ped2. A Multivariate Gaussian Fully Convolutional Adversarial Autoencoder in \cite{li2019video} scored 91.6\% AUC on UCSD Ped2. The spatiotemporal 3D Convolutional Auto-Encoder (ST-3DCAE) in \cite{hu2022video} achieved 85.3\% on UCSD Ped2 and 75.8\% on UCF-Crime. OF-ConvAE-LSTM in \cite{duman2019anomaly} reached 92.9\% on UCSD Ped2 and 89.5\% on Avenue. Finally, \cite{mishra2024anomaly} achieved 86.4\% on UCSD Peds1 and 88.9\% on Avenue using a deep autoencoder with regularity-based thresholding.

\medskip
While deep learning models have significantly advanced video anomaly detection, statistical methods, particularly functional data analysis (FDA), remain underexplored in this area. FDA offers unique advantages by treating each frame’s reconstruction error as a continuous, multivariate function over time, enabling the capture of both temporal and spatial dependencies and allowing for a more effective distinction between typical fluctuations and true anomalies by analyzing the shape and magnitude of deviations \cite{dai2018multivariate}. Despite these benefits, only a few studies have investigated statistical approaches in video anomaly detection. For example, Raymaekers et al. introduced a measure of directional outlyingness specifically applied to image and video data \cite{rousseeuw2018measure}, using a statistical method for anomaly detection based on directional outlyingness. However, this approach has only been validated on images and non-crowded video sequences, emphasizing the need for further testing in more complex, crowded environments. Another statistical approach for anomaly detection in high-dimensional data is the Depthgram method by Aleman et al. \cite{aleman2022depthgram}, designed for functional data visualization in fMRI. Depthgram uses depth-based 2D representations to identify outliers, variability, and sample composition, supporting the exploration of neuroscientific patterns across individuals and brain regions. However, it is limited to static images, such as fMRI data, and has not been explored for video applications.

\medskip
This study explores video anomaly detection in crowded scenes within a Functional Data Analysis framework, offering a statistical perspective to enhance the detection process. Many statistical video anomaly detection methods \cite{naji2022spatio,yang2025follow,singh2023stemgan,le2023attention,cao2024context,park2022fastano,ristea2024self,hong2024making,liu2023diversity} often rely on thresholds determined in a non-automatic and non-systematic manner, focusing primarily on anomalies in the mean or variance of reconstructed residuals. To mitigate these limitations, this work explores the application of the Magnitude-Shape (MS) Plot, a Multivariate Functional Data Visualization and Outlier Detection approach, for video anomaly detection~\cite{dai2018multivariate}. The MS-Plot provides a statistically principled and interpretable framework by treating reconstruction errors as multivariate functional data. It effectively captures both magnitude (amplitude deviations) and shape (pattern deviations), allowing for the detection of subtle anomalies and complex patterns indicative of abnormal behavior. By simultaneously monitoring amplitude and shape deviations, this approach provides a comprehensive statistical methodology to analyze and understand reconstruction errors in video anomaly detection. In addition, the approach trains solely on normal data, making it effective in scenarios with scarcely labeled anomalies by learning typical patterns and identifying deviations during evaluation. To illustrate, this study investigates two reconstruction-based models: a simple autoencoder and the advanced MAMA autoencoder \cite{hong2024making}. These models learn and reconstruct normal behavior from anomaly-free datasets, producing low reconstruction errors for normal frames and significantly higher errors for anomalous frames during testing. The integration of the MS-Plot enhances this approach by jointly analyzing magnitude and shape deviations, offering a comprehensive framework for anomaly detection. This approach was evaluated on two publicly available datasets, UCSD Ped2 and CUHK Avenue, demonstrating superior performance compared to traditional univariate functional detectors,  including Functional Boxplot (FBplot)~\cite{sun2011functional}, Total Variation Depth with Modified Shape Similarity (TVDMSS), Extremal Depth (ED)~\cite{narisetty2016extremal}, and Outliergram (OG)~\cite{arribas2014shape}.  Unlike these univariate approaches, which focus on either shape or mean deviations, MS-Plot captures complex, multivariate outliers, proving more sensitive to both subtle and significant anomalies in video data. Also, the proposed approach surpassing state-of-the-art techniques.

\medskip 
The remaining sections of this paper are organized as follows.  Section~\ref{sec2} provides a brief overview of the MS-Plot approach and its application in anomaly detection, along with a recap of the simple autoencoder and other variants utilized in this study. Section~\ref{sec3} details the proposed approach based on MS-Plot. Section~\ref{sec4} discusses the datasets employed and presents the evaluation results that validate the effectiveness of our method. Finally, Section~\ref{sec5} concludes the study.

\section{Materials and methods}\label{sec2}
\subsection{Problem statement}

Video anomaly detection aims to identify frames with abnormal behavior by learning normal patterns during training and detecting deviations during testing. Frame-level detection evaluates each frame for anomalies, such as unexpected objects, irregular behaviors, or environmental changes, without relying on object- or scene-based approaches. Let \( X \) denote a video consisting of multiple frames \( X = \{ X_1, X_2, \ldots, X_T \} \), where \( T \) represents the total number of frames. Each frame \( X_t \) at time \( t \) is represented as:
\begin{equation}
X_t = \{ p_{ij}(t) \mid 1 \leq i \leq H, \; 1 \leq j \leq W \},
\end{equation}
\noindent where \( H \) and \( W \) are the height and width of the frame, and \( p_{ij}(t) \) denotes the intensity of the pixel located at \( (i, j) \) in frame \( t \).

The goal is to develop a model \( f \) that assigns a binary label \( y_t \) to each frame \( X_t \) such that:

\begin{equation}
y_t = f(X_t) \quad \text{where} \quad y_t \in \{ 0, 1 \}
\end{equation}

Here, \( y_t = 0 \) indicates a normal frame, while \( y_t = 1 \) indicates an anomalous frame. This detection is performed at the frame level.

\medskip
Frame-level anomaly detection in videos presents challenges such as diverse anomaly types and limited labeled data. Anomalies range from sudden lighting changes to unexpected behaviors, complicating model generalization. The scarcity of labeled data necessitates unsupervised or semi-supervised methods focused on normal patterns. This study employs autoencoder models combined with the MS-Plot to enhance video anomaly detection. The following sections elaborate on the MS-Plot and its role in this process.

\subsection{MS-Plot for Functional Data Outlier Detection}
The MS-Plot, introduced by Dai and Genton (2018) \cite{dai2018multivariate}, detects outliers in functional data by evaluating magnitude and shape deviations from a central region. Extending the concept of functional directional outlyingness, it measures both the extent and direction of deviations. This measure allows the MS-Plot to effectively capture two components of outlyingness: magnitude (distance from the central region) and shape (direction of deviation), making it suitable for both univariate and multivariate functional data.

\begin{itemize}
    \item Mean Directional Outlyingness (MO): This quantifies the average magnitude of a function’s deviation from the central region across its domain, effectively identifying anomalies in overall amplitude or intensity. For a function \( X(t) \) with distribution \( F_X \), it is defined as \cite{dai2018multivariate}:
    \begin{equation}
    MO(X, F_X) = \int_I O(X(t), F_{X(t)}) w(t) \, dt,
    \end{equation}
    where \( O(X(t), F_{X(t)}) \) represents the directional outlyingness at each point \( t \) in the domain \( I \), and \( w(t) \) is a weight function.
    
    \item Variation of Directional Outlyingness (VO): This measures the variability in directional outlyingness across the domain, capturing deviations in the shape or structure of the function. By assessing this variability, VO enables the detection of shape anomalies or irregular patterns within the data \cite{dai2018multivariate}.
    \begin{equation}
    VO(X, F_X) = \int_I \|O(X(t), F_{X(t)}) - MO(X, F_X)\|^2 w(t) \, dt.
    \end{equation}
\end{itemize}

Functional Directional Outlyingness (FO) quantifies the overall outlyingness of a function, defined as \cite{dai2018multivariate}:
\begin{equation}\label{Eq:FO}
FO(X, F_X) = \|MO(X, F_X)\|^2 + VO(X, F_X).
\end{equation}

In Equation (\ref{Eq:FO}), FO combines magnitude outlyingness \( \|MO\| \), representing deviation extent, and shape outlyingness \( VO \), reflecting structural variability. This decomposition enhances the ability to quantify centrality and identify abnormal patterns in functional data.

\medskip 
The MS-Plot technique classifies observations as normal or anomalous by applying a threshold to the FO measure, which combines magnitude and shape deviations. Dai and Genton~\cite{dai2018multivariate} proposed calculating FO using MO and VO. Assuming normal data distribution and using random projection point-wise depth, they compute functional directional outlyingness. The method utilizes the squared robust Mahalanobis distance (SRMD) for \((MO, VO)^T\), where the covariance matrix is determined using the Minimum Covariance Determinant (MCD) algorithm~\cite{rousseeuw1999fast}. The SRMD’s tail distribution is approximated with a Fisher's F distribution~\cite{hardin2005distribution}, and curves with SRMD values exceeding the threshold are flagged as outliers.

\subsection{Autoencoder-Based Video Anomaly Detection}
An autoencoder is a neural network framework that learns to compress input data into a latent representation and then reconstruct it back to its original form~\cite{bengio2013representation}. This process is facilitated by two key components: an encoder, which reduces the data's dimensionality, and a decoder, which restores it~\cite{hinton2006reducing}.

\begin{itemize}
\item \textbf{Encoder:} transforms the input data into a compact latent space, preserving critical features and filtering out noise and unnecessary details.
\item \textbf{Decoder:} Reconstructs the input data from the latent representation, aiming to reduce the reconstruction error by aligning the output closely with the original input.
\end{itemize}

\medskip 
Let \( \mathbf{X} \) represent an input frame. The encoder function \( E \) transforms \( \mathbf{X} \) into a latent representation \( \mathbf{Z} \), and the decoder function \( D \) reconstructs the original frame as \( \mathbf{\hat{X}} \):
\begin{equation}
\mathbf{Z} = E(\mathbf{X}), \quad \mathbf{\hat{X}} = D(\mathbf{Z}).
\end{equation}

During training, the autoencoder is optimized to minimize a reconstruction loss, commonly the Mean Squared Error (MSE) between the input \( \mathbf{X} \) and its reconstruction \( \mathbf{\hat{X}} \):
\begin{equation}
L(\mathbf{X}, \mathbf{\hat{X}}) = \frac{1}{N} \sum_{i=1}^{N} (\mathbf{X}_i - \mathbf{\hat{X}}_i)^2,
\end{equation}
\noindent where \( N \) is the total number of pixels in the frame.

\begin{figure}[h!]
    \centering
    \includegraphics[width=12cm]{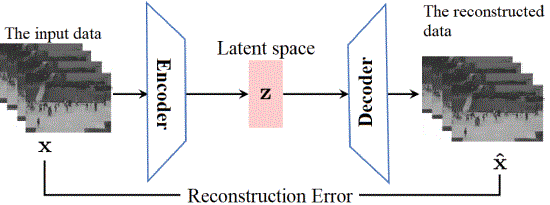}
    \caption{Basic Autoencoder for Video Anomaly Detection.}
    \label{AE_VAD}
\end{figure}

Figure~\ref{AE_VAD} presents a flowchart of an autoencoder for video anomaly detection. Trained solely on normal frames, the autoencoder minimizes reconstruction errors for normal patterns. Anomalous frames deviating from these patterns result in higher reconstruction errors, forming the basis for anomaly detection.

\medskip 
The process of using an autoencoder for video anomaly detection is summarized as follows.
\begin{enumerate}
    \item \textbf{Training Phase:} The autoencoder is trained on a dataset of normal video frames, optimizing it to reconstruct typical, expected frames.
    \item \textbf{Testing Phase:} For each test frame \( \mathbf{X}_t \), the autoencoder reconstructs the frame as \( \mathbf{\hat{X}}_t \) and computes the reconstruction error:
    \begin{equation}
    e_t = \| \mathbf{X}_t - \mathbf{\hat{X}}_t \|,
    \end{equation}
  \noindent  where \( e_t \) denotes the overall reconstruction error for frame \( \mathbf{X}_t \).
    \item \textbf{Anomaly Detection:} A frame is identified as an anomaly when the reconstruction error \( e_t \) surpasses a specified threshold  \( \tau \).
    \begin{equation}
    y_t = \begin{cases}
        1 & \text{if } e_t > \tau \\
        0 & \text{otherwise}.
    \end{cases}
    \end{equation}
    Here, \( y_t = 1 \) indicates an anomalous frame, while \( y_t = 0 \) indicates a normal frame.
\end{enumerate}

\medskip 
Autoencoders effectively detect video anomalies by learning compact representations of normal patterns, highlighting subtle deviations through reconstruction errors~\cite{sabokrou2016video}. Trained exclusively on normal data, they excel in unsupervised settings, making them suitable for applications like surveillance and traffic monitoring~\cite{duong2023deep}. Their ability to flag deviations without labeled anomalies is especially valuable in scenarios with limited anomaly data~\cite{gong2019memorizing}. Various enhancements have been proposed to improve their detection capabilities~\cite{hasan2016learning,zhao2017spatio,hong2024making,liu2023diversity}.

\section{MS-Plot for Video Anomaly Detection with Reconstruction Models}\label{sec3}
This study introduces a framework that integrates the MS-Plot with reconstruction-based models for anomaly detection. Residuals (reconstruction errors) generated by these models are treated as multivariate functional data and analyzed using the MS-Plot to identify anomalies. Frame-level anomaly detection involves learning normal behavior patterns, calculating reconstruction errors, and applying the MS-Plot for multivariate functional outlier analysis. This enhances detection accuracy and versatility, enabling the identification of diverse anomalies in video sequences. For simplicity, the MS-Plot framework is demonstrated with an autoencoder model, but it can be extended to other reconstruction-based models.

\medskip
This approach consists of three steps: training a reconstruction-based model on anomaly-free data to learn normal patterns, generating residuals by comparing input and reconstructed frames during testing, and analyzing these residuals with the MS-Plot to detect anomalies based on magnitude and shape outlyingness. This integration of reconstruction learning and multivariate analysis ensures effective anomaly detection.

\subsection{Training Phase - Learning Normal Behavior}
During training, only normal frames are provided to the model. The model is typically an autoencoder comprising an encoder function \( E \) and a decoder function \( D \). The model aims to reconstruct the input frame \( \mathbf{X}_t \) as \( \mathbf{\hat{X}}_t \):
\begin{equation}
\mathbf{\hat{X}}_t = D(E(\mathbf{X}_t))
\end{equation}

The autoencoder is trained to minimize a reconstruction loss function \( L(\mathbf{X}_t, \mathbf{\hat{X}}_t) \), which measures the difference between the original frame \( \mathbf{X}_t \) and the reconstructed frame \( \mathbf{\hat{X}}_t \). A common choice for \( L \) is the mean squared error (MSE):
\begin{equation}
L(\mathbf{X}_t, \mathbf{\hat{X}}_t) = \frac{1}{HW} \sum_{i=1}^{H} \sum_{j=1}^{W} (p_{ij}(t) - \hat{p}_{ij}(t))^2
\end{equation}
where \( \hat{p}_{ij}(t) \) represents the intensity of the reconstructed pixel at location \( (i, j) \).

\subsection{Testing Phase - Residual Generation and Anomaly Detection}
In the testing phase, the model receives frames from unseen video sequences. For each frame \( \mathbf{X}_t \), it computes the reconstruction \( \mathbf{\hat{X}}_t \) and the corresponding reconstruction error matrix:
\begin{equation}
\mathbf{E}_t = \{ e_{ij}(t) \mid 1 \leq i \leq H, \; 1 \leq j \leq W \}
\end{equation}
\noindent where \( e_{ij}(t) = | p_{ij}(t) - \hat{p}_{ij}(t) | \) denotes the absolute difference between the original and reconstructed pixel intensities.

\medskip 
To enhance anomaly detection, the reconstruction error matrices \( \mathbf{E}_t \) are treated as multivariate functional data. For each frame \( \mathbf{X}_t \), the MS-Plot calculates MO and VO to detect deviations from normal patterns. The directional outlyingness \( O(\mathbf{E}_t, F) \) quantifies the extent to which the error matrix \( \mathbf{E}_t \) deviates from a reference distribution \( F \):
\begin{equation}
MO(\mathbf{E}_t, F) = \int_I O(\mathbf{E}_t, F_{\mathbf{E}_t}) \, w(t) \, dt, \quad VO(\mathbf{E}_t, F) = \int_I \| O(\mathbf{E}_t, F_{\mathbf{E}_t}) - MO(\mathbf{E}_t, F) \|^2 w(t) \, dt.
\end{equation}

The MS-Plot visualizes \( |MO| \) against \( VO \) to distinguish normal from anomalous frames, with the relationship defined as:
\begin{equation}
FO = \|MO\|^2 + VO.
\end{equation}

\medskip 
The MS-Plot applies a threshold on functional outlyingness \( FO \) to classify each frame \( \mathbf{X}_t \) as normal or anomalous. The decision rule is:
\begin{equation}
y_t = \begin{cases}
    1 & \text{if } FO(\mathbf{E}_t) > \tau_{FO} \\
    0 & \text{otherwise}.
\end{cases}
\end{equation}

\medskip
Algorithm~\ref{alg:ms_plot_anomaly_detection} summarizes the steps for using the MS-Plot to analyze residuals and detect frame-level anomalies.
\begin{algorithm}[h!]
\caption{MS-Plot framework for anomaly detection based on reconstruction residuals.}
\label{alg:ms_plot_anomaly_detection}
\begin{algorithmic}[1]
\REQUIRE Functional data (reconstruction error matrices) $\{\mathbf{E}_1, \mathbf{E}_2, \dots, \mathbf{E}_n\}$, Reference distribution $F_{\mathbf{E}}$, Threshold $\tau_{FO}$
\ENSURE Anomaly detection result for each $\mathbf{E}_i$ (0: normal, 1: anomaly)

\STATE Compute the reference distribution $F_{\mathbf{E}(t)}$ (e.g., median function) at each time point $t$

\FOR{each error matrix $\mathbf{E}_i$}
    \STATE \textbf{Compute directional outlyingness} $O(\mathbf{E}_i(t), F_{\mathbf{E}(t)})$ at each time point $t$:
    \[
    O(\mathbf{E}_i(t), F_{\mathbf{E}(t)}) = (\mathbf{E}_i(t) - F_{\mathbf{E}(t)}) \times \text{Sign}(\mathbf{E}_i(t) - F_{\mathbf{E}(t)})
    \]
    \STATE \textbf{Calculate Mean Directional Outlyingness (MO) for $\mathbf{E}_i$:}
    \[
    MO(\mathbf{E}_i) = \frac{1}{T} \sum_{t=1}^{T} O(\mathbf{E}_i(t), F_{\mathbf{E}(t)})
    \]
    \STATE \textbf{Calculate Variation of Directional Outlyingness (VO) for $\mathbf{E}_i$:}
    \[
    VO(\mathbf{E}_i) = \frac{1}{T} \sum_{t=1}^{T} \left( O(\mathbf{E}_i(t), F_{\mathbf{E}(t)}) - MO(\mathbf{E}_i) \right)^2
    \]
    \STATE \textbf{Calculate Functional Outlyingness (FO) for $\mathbf{E}_i$:}
    \[
    FO(\mathbf{E}_i) = \|MO(\mathbf{E}_i)\|^2 + VO(\mathbf{E}_i)
    \]
\ENDFOR

\STATE \textbf{Construct MS-Plot:} Plot $|MO(\mathbf{E}_i)|$ on the x-axis and $VO(\mathbf{E}_i)$ on the y-axis for each error matrix $\mathbf{E}_i$

\STATE \textbf{Set Threshold and Detect Outliers:}
\FOR{each error matrix $\mathbf{E}_i$}
    \IF{$FO(\mathbf{E}_i) > \tau_{FO}$}
        \STATE Classify $\mathbf{E}_i$ as anomaly $(y_i = 1)$
    \ELSE
        \STATE Classify $\mathbf{E}_i$ as normal $(y_i = 0)$
    \ENDIF
\ENDFOR

\RETURN Anomaly labels $\{y_1, y_2, \dots, y_n\}$ for all error matrices
\end{algorithmic}
\end{algorithm}

\subsection{Evaluation Metrics for Anomaly Detection in Videos}
In video anomaly detection, evaluation metrics like True Positive Rate (TPR), False Positive Rate (FPR), Precision, F1 Score, and Accuracy are essential for assessing the effectiveness of identifying anomalies while minimizing false detections.  These metrics quantify detection effectiveness by comparing model predictions against ground truth labels. The mathematical definitions of these metrics are summarized in Table~\ref{tab:evaluation_metrics}, where \( \text{TP} \), \( \text{FP} \), \( \text{TN} \), and 
\( \text{FN} \) represent True Positives, False Positives, True Negatives, and False Negatives, respectively. 
\begin{table}[h!]
\centering
\caption{Summary of Evaluation Metrics for Video Anomaly Detection.}
\label{tab:evaluation_metrics}
\begin{tabular}{l|l}
\hline
\textbf{Metric}        & \textbf{Equation} \\
\hline
True Positive Rate (TPR) & \( \text{TPR} = \frac{\text{TP}}{\text{TP} + \text{FN}} \) \\
\hline
False Positive Rate (FPR) & \( \text{FPR} = \frac{\text{FP}}{\text{FP} + \text{TN}} \) \\
\hline
Precision                & \( \text{Precision} = \frac{\text{TP}}{\text{TP} + \text{FP}} \) \\
\hline
F1 Score                 & \( \text{F1} = 2 \cdot \frac{\text{Precision} \cdot \text{Recall}}{\text{Precision} + \text{Recall}} \) \\
\hline
Accuracy                 & \( \text{Accuracy} = \frac{\text{TP} + \text{TN}}{\text{TP} + \text{FP} + \text{TN} + \text{FN}} \) \\
\hline
\end{tabular}
\end{table}

TPR measures the proportion of actual anomalies correctly identified, while FPR indicates the rate of normal frames misclassified as anomalies, a critical factor in reducing false alarms in video data. Precision evaluates the accuracy of predicted anomalies, reflecting the proportion of correctly identified anomalies among all flagged instances. Accuracy captures the overall proportion of correctly classified frames, encompassing both normal and anomalous ones. Finally, the F1-score provides a balanced assessment by combining Precision and Recall, particularly useful in scenarios with class imbalance.

\newpage
\section{Results and discussion}\label{sec4}

\subsection{Data discription}
In this section, the investigated anomaly detection methods are tested on two widely-used benchmark datasets: the UCSD Pedestrian (Ped2) dataset \cite{mahadevan2010anomaly} and the CUHK Avenue dataset \cite{lu2013abnormal}, summarized in Table~\ref{tab:datasets}.
\begin{table}[h!]
\centering
\caption{Details of the UCSD Ped2 and CUHK Avenue Datasets.}
\label{tab:datasets}
\begin{adjustbox}{width=1\textwidth}
\begin{tabular}{l|c|c|c|c}
\hline
\textbf{Dataset} & \textbf{Total Frames} & \textbf{Training Frames} & \textbf{Testing Frames} & \textbf{Resolution (pixels)} \\
\hline
UCSD Ped2 & 4,560 & 2,550 & 2,010 & 360×240 \\
CUHK Avenue & 30,652 & 15,328 & 15,324 & 640×360 \\
\hline
\end{tabular}
\end{adjustbox}
\end{table}

\begin{figure}[h!]
    \centering
    \includegraphics[width=11cm]{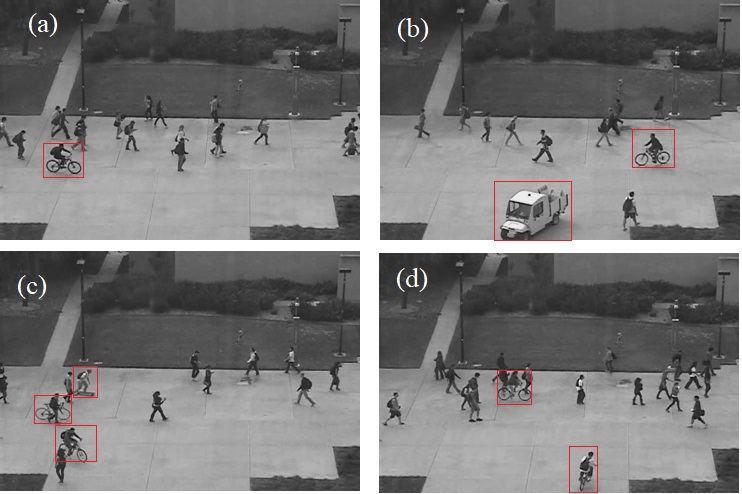}
    \caption{Sample frames from the UCSD Ped2 dataset showing anomalies: (a) Cyclist, (b) Van and cyclist, (c) Cyclist and skater, and (d) two cyclists. Red boxes highlight anomalous regions.}
    \label{ExamplesPed2}
\end{figure}

\noindent\textbf{UCSD Ped2 dataset:} The UCSD Ped2 dataset \cite{mahadevan2010anomaly} is a widely-used benchmark for video anomaly detection, particularly in surveillance contexts. It captures video sequences of a pedestrian-only walkway, where anomalies such as bicycles, skateboards, and vehicles disrupt typical pedestrian activity. The dataset is divided into a training set with 16 clips (2,550 frames) containing only normal pedestrian behavior and a testing set with 12 clips (2,010 frames) featuring both normal events and anomalies. This structure allows models to learn typical patterns during training and evaluate their ability to detect anomalies. Figure~\ref{ExamplesPed2} illustrates examples from the dataset, highlighting various anomalies, including cyclists, vehicles, and skaters, which challenge models to accurately differentiate normal and abnormal events.
\medskip

\medskip 
The UCSD Ped2 dataset is recorded at a resolution of 240×360 pixels with a frame rate of 10 frames per second (fps). It includes frame- and pixel-level annotations in the test set, marking anomalies, anomalies timing and location, and facilitating detailed performance evaluation. Frame-level annotations assess a model's ability to detect anomalies within a frame, while pixel-level annotations enable precise localization of anomalous events. Its controlled environment and clear distinction between normal and anomalous events make it a widely accepted standard for evaluating video anomaly detection models. This dataset is essential for benchmarking new methods, assessing model accuracy in anomaly detection, and comparing performance across established metrics.  Table \ref{tab:anomalous_frames} outlines the anomalous frames and total video lengths in the UCSD Ped2 dataset, focusing on videos with both normal and anomalous events for MS-Plot analysis. 
\begin{table}[h!]
\centering
\caption{Anomalous Frame Ranges and Total Frame Count for Each Test Video in the UCSD Ped2 Dataset.}
\label{tab:anomalous_frames}
\begin{tabular}{l|c|c}
\hline
\textbf{Test Video} & \textbf{Anomalous Frame Range} & \textbf{Total Frames} \\
\hline
Video 1    & 61–180                & 180          \\
Video 2    & 95–180                & 180          \\
Video 3    & 1–146                 & 150          \\
Video 4    & 31–180                & 180          \\
Video 5    & 1–129                 & 150          \\
Video 6    & 1–159                 & 180          \\
Video 7    & 46–180                & 180          \\
Video 8    & 1–180                 & 180          \\
Video 9    & 1–120                 & 120          \\
Video 10   & 1–150                 & 150          \\
Video 11   & 1–180                 & 180          \\
Video 12   & 88–180                & 180          \\
\hline
\end{tabular}
\end{table}

\begin{figure}[h!]
    \centering
    \includegraphics[width=5.2cm]{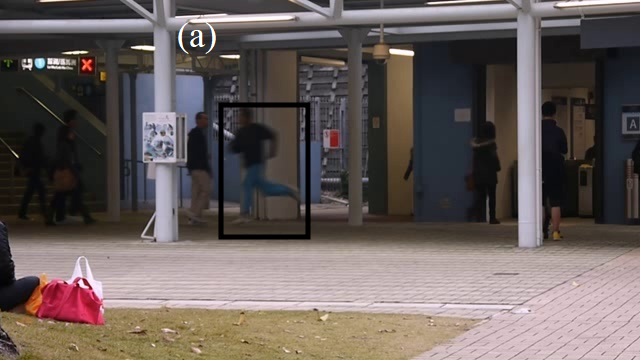}
      \includegraphics[width=5.2cm]{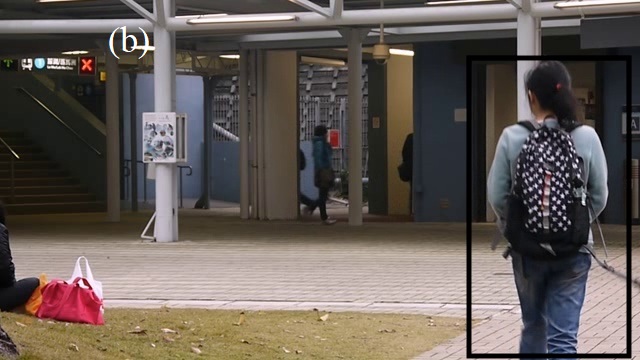}
        \includegraphics[width=5.2cm]{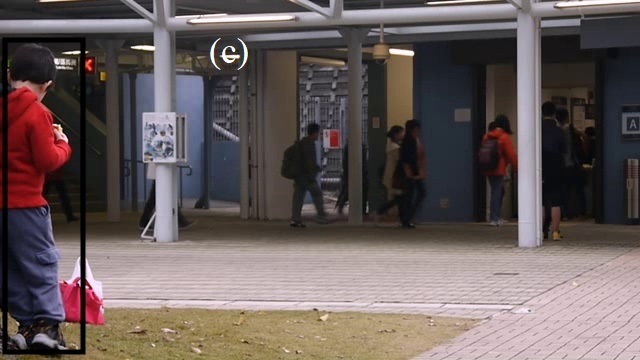}
    \caption{Sample frames from the CUHK Avenue dataset showing anomalies: (a) running, (b) walking in unconventional directions, and (c) loitering.}
    \label{ExamplesCUHK}
\end{figure}

\medskip 
\noindent\textbf{CUHK Avenue dataset:}
The CUHK Avenue dataset \cite{lu2013abnormal} is a widely used benchmark for video anomaly detection. Captured by a fixed camera overlooking an avenue at the Chinese University of Hong Kong, it offers consistent scene composition, varied lighting conditions, and interactions among multiple pedestrians. The dataset contains 37 video sequences, recorded at 25 frames per second (fps) with a resolution of 640 × 360 pixels, comprising approximately 30,652 frames. It includes 16 training videos featuring normal pedestrian activities, such as walking and entering or exiting the scene, and 21 testing videos containing both normal and abnormal events. Abnormal behaviors include running, throwing objects, loitering, walking in unusual directions, and abandoning items. Figure~\ref{ExamplesCUHK} presents examples from the CUHK Avenue dataset, showcasing a range of anomalies, such as running, walking in unconventional directions, and loitering.

%\newpage
\subsection{Detection results using UCSD2 Ped2 dataset}
In the first experiment, the potential of the MS-Plot technique as an effective detection framework is demonstrated by employing three distinct reconstruction-based models: a simple autoencoder and the MAMA-based model~\cite{hong2024making}. Each model is trained exclusively on anomaly-free data to establish a baseline representation of normal behavior. The autoencoder architecture used in this study features an encoder with three convolutional blocks, each comprising a 3×3 Conv2D layer with ReLU activation followed by a 2×2 MaxPooling layer. The encoder’s convolutional layers include 32, 16, and 8 filters. The decoder mirrors this structure, using 3×3 Conv2D layers with ReLU activation and 2×2 upsampling, concluding with a 3×3 Conv2D layer (three filters) and a Sigmoid activation function for output generation. Input images are resized to 64×64 pixels, normalized to [0, 1], and processed in batches of 32. The model is trained for 50 epochs using the Adam optimizer and a Mean Squared Error (MSE) loss function to measure reconstruction error.

The second approach combines the MAMA reconstruction-based model~\cite{hong2024making} with the MS-Plot for anomaly detection. The MAMA model operates in two stages: the F2LM generator processes video frames through three parallel streams—raw frames, semantic labels (DeepLabv3), and motion data (FlowNet2)—using a Feature Transform Convolutional (FTC) block to generate high-quality reconstructions for normal events and degraded outputs for anomalies. In the second stage, the Destroyer network enhances these distinctions by suppressing low-quality regions in the F2LM output. To enhance temporal context and motion evolution, the model processes five consecutive frames at each time point. The residuals generated by the two considered reconstruction-based models are analyzed using the MS-Plot, which takes the residual matrix as input and evaluates anomalies through a multidimensional assessment of magnitude and shape outlyingness. By treating the residual matrix as functional data, the MS-Plot detects subtle and complex anomalies, leveraging its nuanced analysis of deviations and the models’ ability to handle diverse normal patterns, thereby improving sensitivity and accuracy in video anomaly detection.

\medskip 
Tables \ref{AE:msplot} and \ref{MAMA:msplot} summarize the detection performance of the investigated models on the UCSD Ped2 testing set, using the MS-Plot to analyze reconstruction errors. Table \ref{AE:msplot} highlights the results for the AE-MS-Plot method, demonstrating varying performance across the testing videos. The method achieves perfect TPR and AUC scores for certain videos (e.g., Videos 9–11), reflecting accurate detection of anomalies. However, its performance falters in scenarios like Video 8, where cyclists and skaters move in different directions, leading to notably low TPR and Accuracy scores. Similarly, Videos 1 and 5, where a bike appears in crowded areas, show higher false positive rates, indicating difficulty in distinguishing anomalies in complex scenes. These results suggest that while the AE-MS-Plot method can effectively detect clear deviations, it faces challenges in scenarios involving subtle anomalies or overlapping movements, emphasizing the need for models capable of handling diverse anomaly patterns.
\begin{table}[h!]
\centering
\caption{Detection performance of AE-MS-Plot on UCSD Ped2 test videos.}\label{AE:msplot}
\begin{tabular}{c|c|c|c|c|c|c}
\hline
Video & TPR & FPR & Accuracy & Precision  &F1-score & AUC \\
\hline
1 & 100.00 & 46.67 & 84.44 & 81.08 & 89.55 & 78.09 \\
2 & 18.60 & 0.00  & 61.11 & 100 & 31.37 & 77.53 \\
3 & 95.89 & 0.00  & 96.00 & 100 & 97.90 & 98.18 \\
4 & 95.33 & 0.00  & 96.11 & 100 & 97.61 & 97.90 \\
5 & 100.00 & 33.33 & 95.33 & 94.85 & 97.36 & 84.66 \\
6 & 98.74 & 9.52  & 97.78 & 98.74 & 98.74 & 95.06 \\
7 & 63.70 & 0.00  & 72.78 & 100 & 77.83 & 83.13 \\
8 & 23.33 & 0.00  & 23.33 & 100 & 37.84 & 63.41 \\
9 & 100 & 0.00  & 100 & 100 & 100 & 100 \\
10 & 100 & 0.00  & 100 & 100 & 100 & 100 \\
11 & 100 & 0.00  & 100 & 100 & 100 & 100 \\
12 & 98.92 & 0.00  & 99.44 & 100 & 99.46 & 99.52 \\
\hline
\end{tabular}
\end{table}

%----------------------------------------------
\begin{table}[h!]
\centering
\caption{Detection performance of MAMA-MS-Plot on UCSD Ped2 test videos.}\label{MAMA:msplot}
\begin{tabular}{c|c|c|c|c|c|c}
\hline
Dataset & TPR  & FPR & Accuracy & Precision & F1-score & AUC \\
\hline
1 & 97.50 & 1.79 & 97.73 & 99.15 & 98.32 & 98.08 \\
2 & 97.67 & 1.11 & 98.30 & 98.82 & 98.25 & 98.43 \\
3 & 98.59 & 0.00 & 98.63 & 100.00 & 99.29 & 99.38 \\
4 & 98.00 & 0.00 & 98.30 & 100.00 & 98.99 & 99.24 \\
5 & 97.60 & 4.76 & 97.26 & 99.19 & 98.39 & 96.60 \\
6 & 97.42 & 4.76 & 97.16 & 99.34 & 98.37 & 96.53 \\
7 & 97.71 & 0.00 & 98.30 & 100.00 & 98.84 & 98.97 \\
8 & 98.86 & 0.00 & 98.86 & 100.00 & 99.43 & 99.66 \\
9 & 97.41 & 0.00 & 97.41 & 100.00 & 98.69 & 98.87 \\
10 & 98.63 & 0.00 & 98.63 & 100.00 & 99.31 & 99.47 \\
11 & 98.30 & 0.00 & 98.30 & 100.00 & 99.14 & 99.32 \\
12 & 95.70 & 1.20 & 97.16 & 98.89 & 97.27 & 97.40 \\
\hline
\end{tabular}
\end{table}

\medskip 
The results in Table \ref{MAMA:msplot} demonstrate the superior performance of the MAMA-MS-Plot method in detecting different types of anomaly in crowded videos within the UCSD Ped2 dataset. The method consistently achieves high TPR, Precision, F1-scores, and AUC values across most videos, demonstrating its capability to identify both clear and subtle anomalies effectively.  Compared to the AE-MS-Plot method, the MAMA-MS-Plot provides enhanced accuracy and maintains low false positive rates, emphasizing its effectiveness in handling diverse testing conditions and crowded environments. This indicates that the sophisticated reconstruction mechanisms in these models generate residuals sensitive enough for the MS-Plot to detect subtle deviations that a simpler autoencoder might miss. Overall, the MS-Plot framework enhances anomaly detection by analyzing magnitude and shape deviations in residuals over time, capturing subtle anomalies, and reducing false positives in complex scenarios.

\newpage
\subsection{Comparison of MS-Plot and univariate functional anomaly detection methods}

The second experiment compares the MS-Plot-based approach with established univariate functional anomaly detection methods, including Functional Boxplot (FBPlot) \cite{sun2011functional}, Total Variation Depth with Modified Shape Similarity (TVDMSS) \cite{huang2019decomposition}, Extremal Depth (ED) \cite{narisetty2016extremal}, and Outliergram (OG) \cite{arribas2014shape}. To apply these methods, residual matrices from the autoencoder models were reshaped into one-dimensional vectors. A brief overview of each method is provided for context before comparing their performance with the MS-Plot.

\begin{itemize}
    \item The FBPlot, introduced by Sun and Genton (2011) ~\cite{sun2011functional}, extends the traditional boxplot for functional data by using depth measures, such as modified band depth (MBD), to detect outliers. It centers on the median curve and constructs an envelope representing the central 50\% of data, identifying outliers as curves that deviate beyond 1.5 times the interquartile range. 
    
    \item The TVDMSS method \cite{huang2019decomposition}, proposed by Huang and Sun (2019), identifies functional outliers using total variation depth (TVD) for magnitude deviations and modified shape similarity (MSS) for shape anomalies. Thresholds on TVD and MSS scores allow detection of outliers in magnitude, shape, or both.
    
    \item The ED \cite{narisetty2016extremal}, proposed by Narisetty and Nair (2016), ranks functional data based on "extremeness" to detect both boundary and central outliers. Curves with ED values below a threshold are flagged as outliers.

    \item The OG \cite{arribas2014shape}, introduced by Arribas-Gil and Romo (2014), identifies shape outliers in functional data using Modified Band Depth (MBD) for centrality and Half-Region Depth (HRD) for spread. Outliers are visualized as points deviating from the main cluster in an MBD-HRD plot, emphasizing unusual shapes.
\end{itemize}

\medskip

Table~\ref{tab:comparison} presents the comparative results of the proposed MS-Plot-based approaches (AE-MS-Plot and MAMA-MS-Plot) against traditional univariate functional anomaly detection methods (FBPlot, TVDMSS, ED, and OG) on the UCSD Ped2 dataset. The analysis highlights notable differences in performance, with MS-Plot-based methods, particularly MAMA-MS-Plot, demonstrating superior results across all evaluated metrics. These findings underscore the effectiveness of MS-Plot in capturing both magnitude and shape deviations, offering a more comprehensive detection framework compared to univariate techniques. For example, MAMA-MS-Plot achieves an AUC of 98.74\%, along with high TPR and precision values (97.90\% and 99.68\%, respectively), demonstrating strong detection accuracy with a minimal false positive rate (FPR of 1.45\%). In comparison, AE-MS-Plot achieves a moderate AUC of 88.3\%, with a TPR of 82.46\%, precision of 94.11\%, and an FPR of 23.48\%.  These results highlight the MS-Plot's effectiveness, with MAMA-MS-Plot excelling in accuracy and sensitivity. In contrast, univariate methods applied to autoencoder residuals perform poorly compared to MS-Plot-based approaches (Table~\ref{tab:comparison}). For instance, AE-FBPlot and AE-TVDMSS achieve an AUC of 52.41\% with a TPR of 27.19\%, reflecting low sensitivity to anomalies. Even the best univariate method, AE-ED, attains an AUC of 60.16\%, far below the performance of MS-Plot-based techniques. Additionally, ED and OG methods applied to vectorized residuals show inconsistent TPR and FPR values, further highlighting their limitations.
\begin{table}[h!]
\centering
\caption{Comparison of average performance: MS-Plot vs. univariate methods on UCSD Ped2 test videos.}
\label{tab:comparison}
\begin{tabular}{l|rrrrrrr}
\toprule
 Approach & TPR & FPR & Accuracy & Precision & F1-score & AUC \\
\midrule
AE-MS-Plot & 82.46 & 23.48 & 81.39 & 94.11 & 87.90 & 88.3 \\
MAMA-MS-Plot  & 97.90 & 1.45 & 98.01 & 99.68 & 98.78 & 98.74 \\
AE-FBPlot & 27.19 & 22.38 & 36.39 & 84.48 & 41.14 & 52.41 \\
AE-TVDMSS & 27.19 & 22.38 & 36.39 & 84.48 & 41.14 & 52.41 \\
AE-ED & 48.40 & 28.07 & 52.50 & 89.09 & 62.72 & 60.16 \\
AE-OG & 0.00 & 0.00 & 17.60 & 0.00 & 0.00 & 50.00 \\
MAMA-FBPlot & 3.09 & 0.00 & 19.98 & 100.00 & 12.76 & 51.54 \\
MAMA-TVDMSS & 3.77 & 0.00 & 20.54 & 100.00 & 15.14 & 51.88 \\
MAMA-ED & 4.69 & 1.75 & 21.00 & 92.68 & 8.93 & 51.47 \\
MAMA-OG & 0.00 & 0.00 & 17.60 & 0.00 & 0.00 & 50.00 \\
DMAD-FBPlot & 0.19 & 0.00 & 17.58 & 100.00 & 0.37 & 50.09 \\
DMAD-TVDMSS & 0.19 & 0.00 & 17.58 & 100.00 & 0.37 & 50.09 \\
DMAD-ED & 11.05 & 17.84 & 23.45 & 74.58 & 19.25 & 46.61 \\
DMAD-OG & 49.81 & 55.56 & 48.88 & 80.94 & 61.67 & 47.13 \\
\bottomrule
\end{tabular}
\end{table}

\medskip
This comparison reveals the superiority of MS-Plot-based methods over univariate approaches. By analyzing both magnitude and shape outlyingness, MS-Plot effectively detects anomalies in functional residuals data. Unlike univariate methods like FBPlot and ED, which rely on single-dimensional measures, MS-Plot captures nuanced deviations in reconstruction errors, critical for frame-level video anomaly detection. Its ability to handle variability in residuals and assess both the level and direction of outlyingness enables the detection of subtle shifts missed by univariate methods, improving overall detection accuracy.

\subsection{MS-Plot Visualization Results based on UCSD Ped2 Testing Set}

\begin{figure}[h!]
    \centering
    \includegraphics[width=6cm]{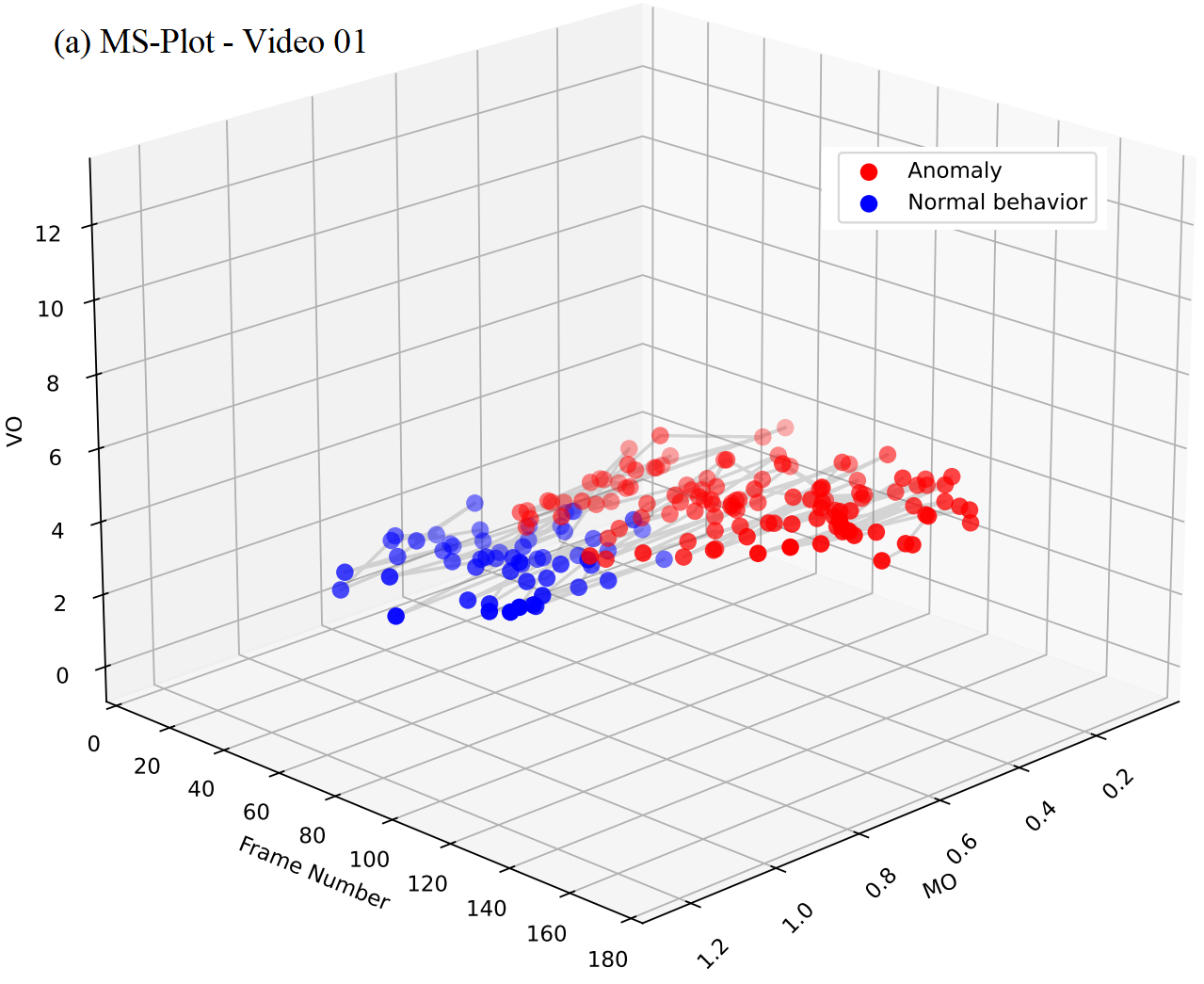}
    \includegraphics[width=6cm]{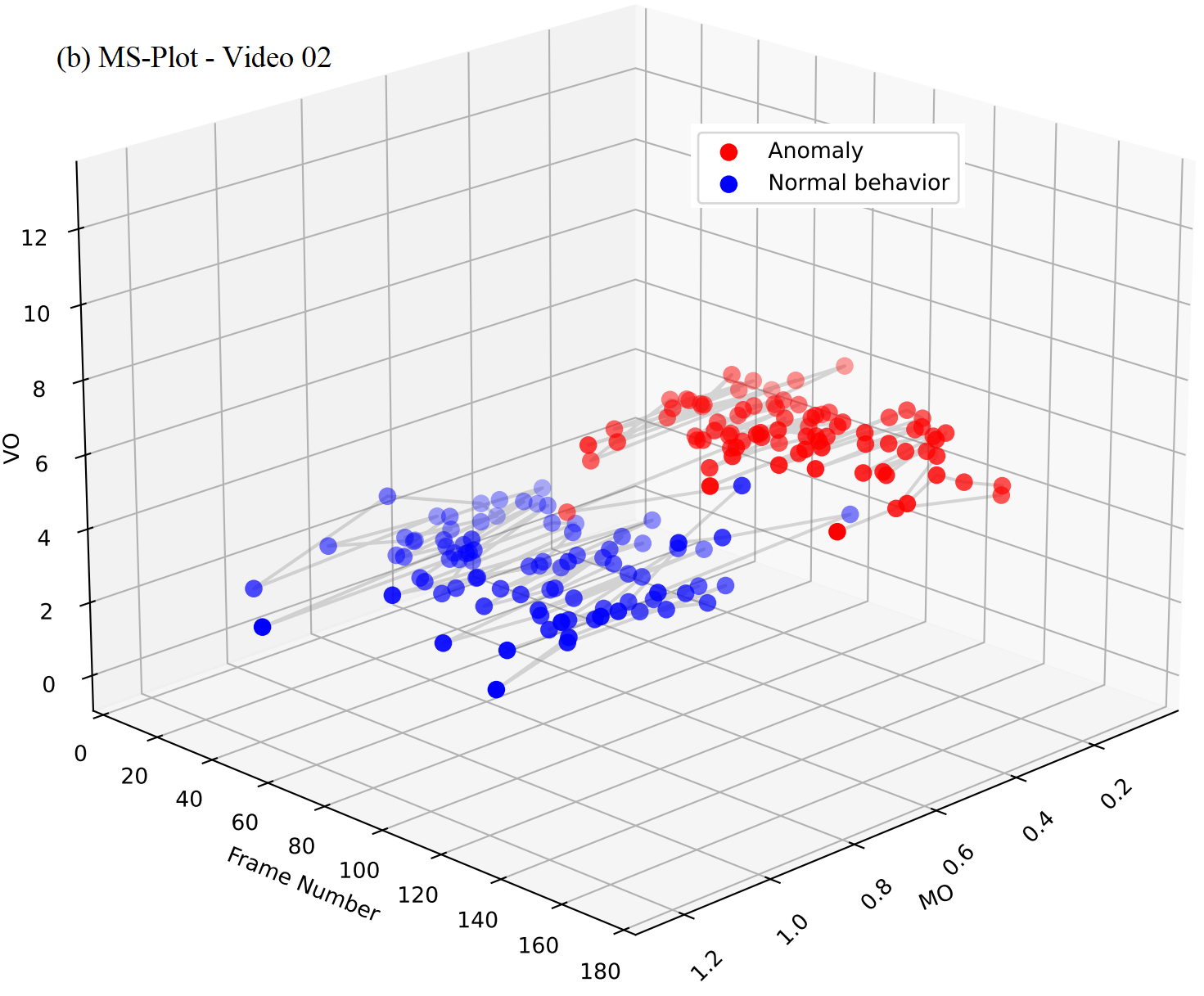}
    \includegraphics[width=6cm]{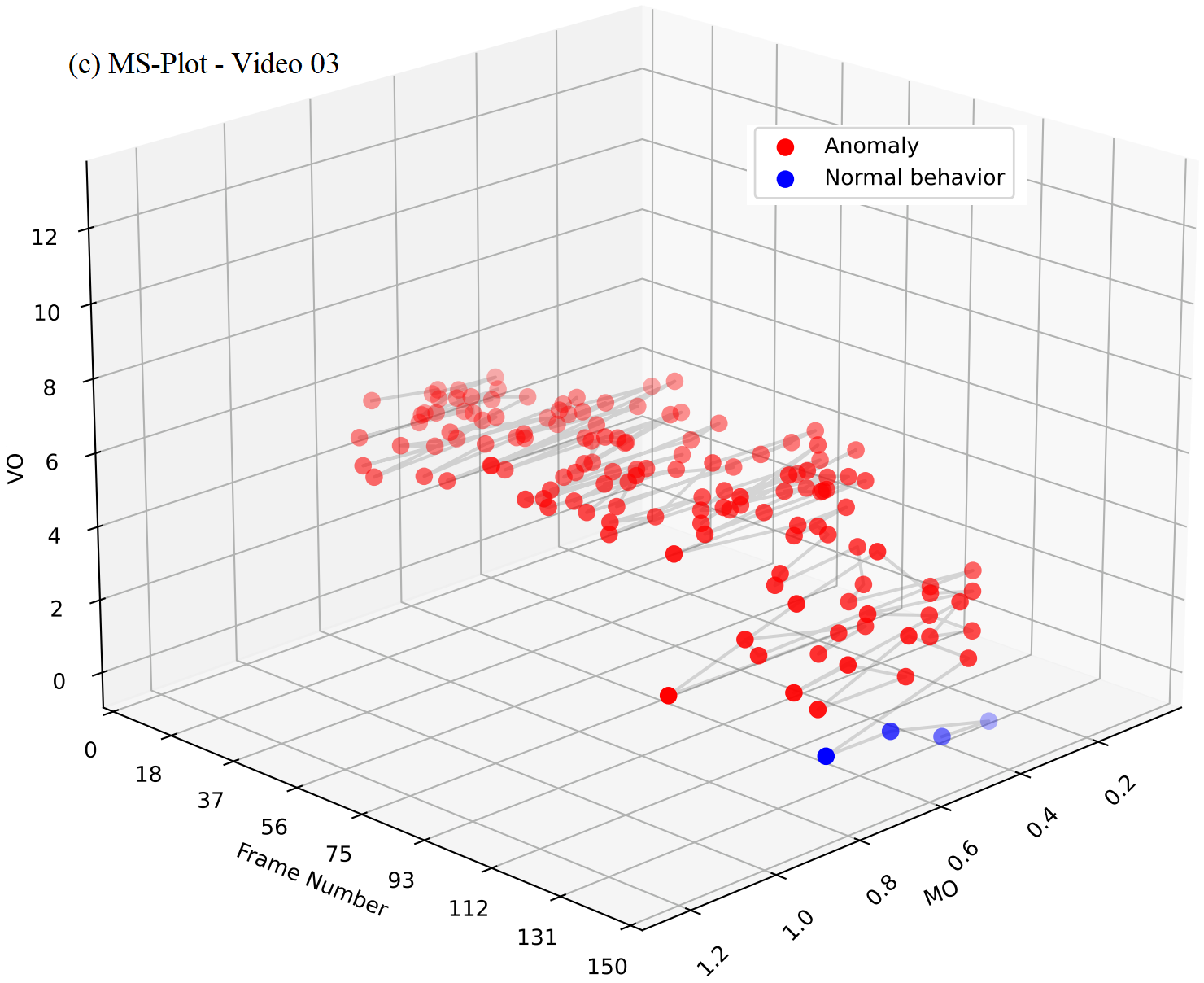}
    \includegraphics[width=6cm]{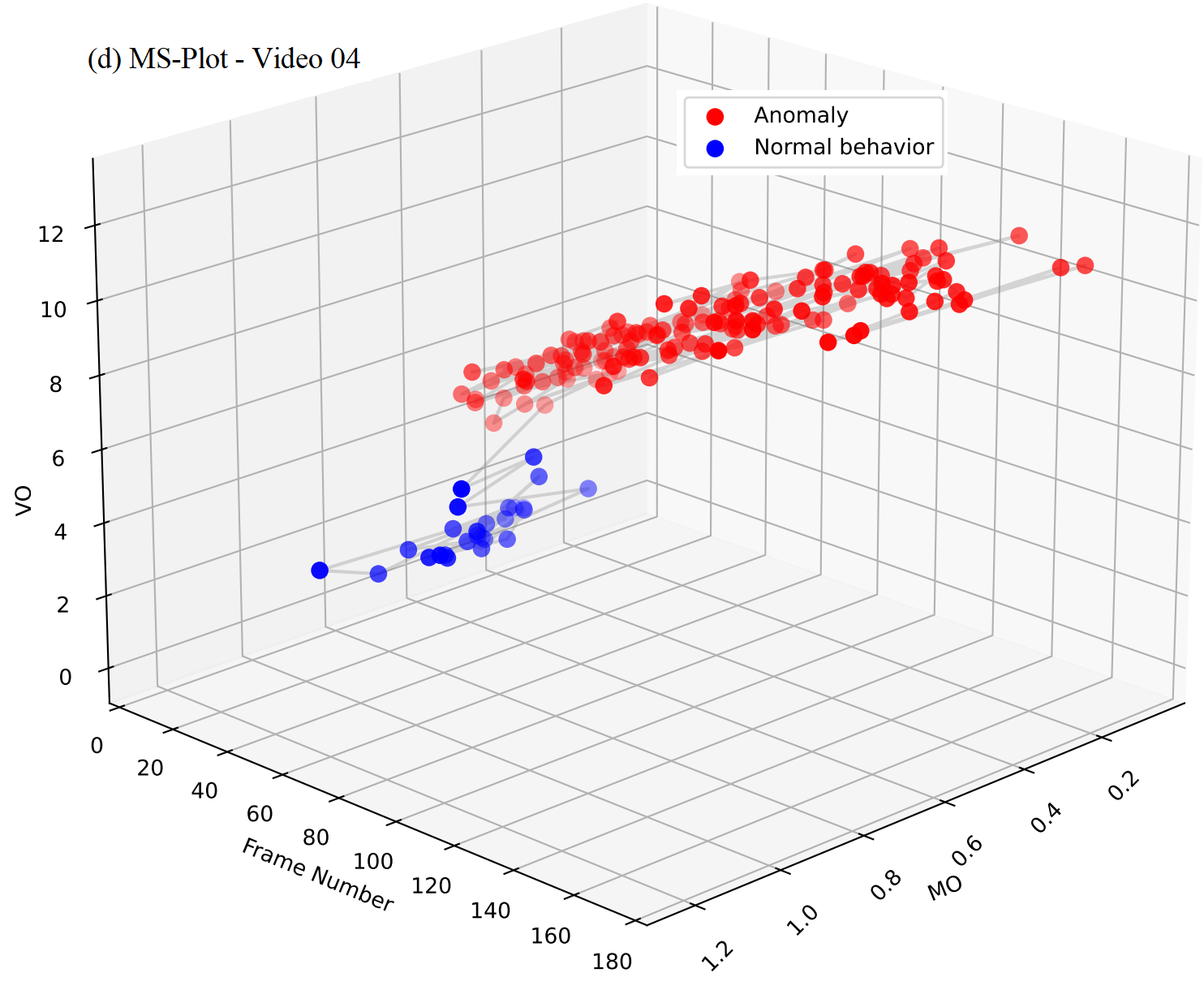}
    \includegraphics[width=6cm]{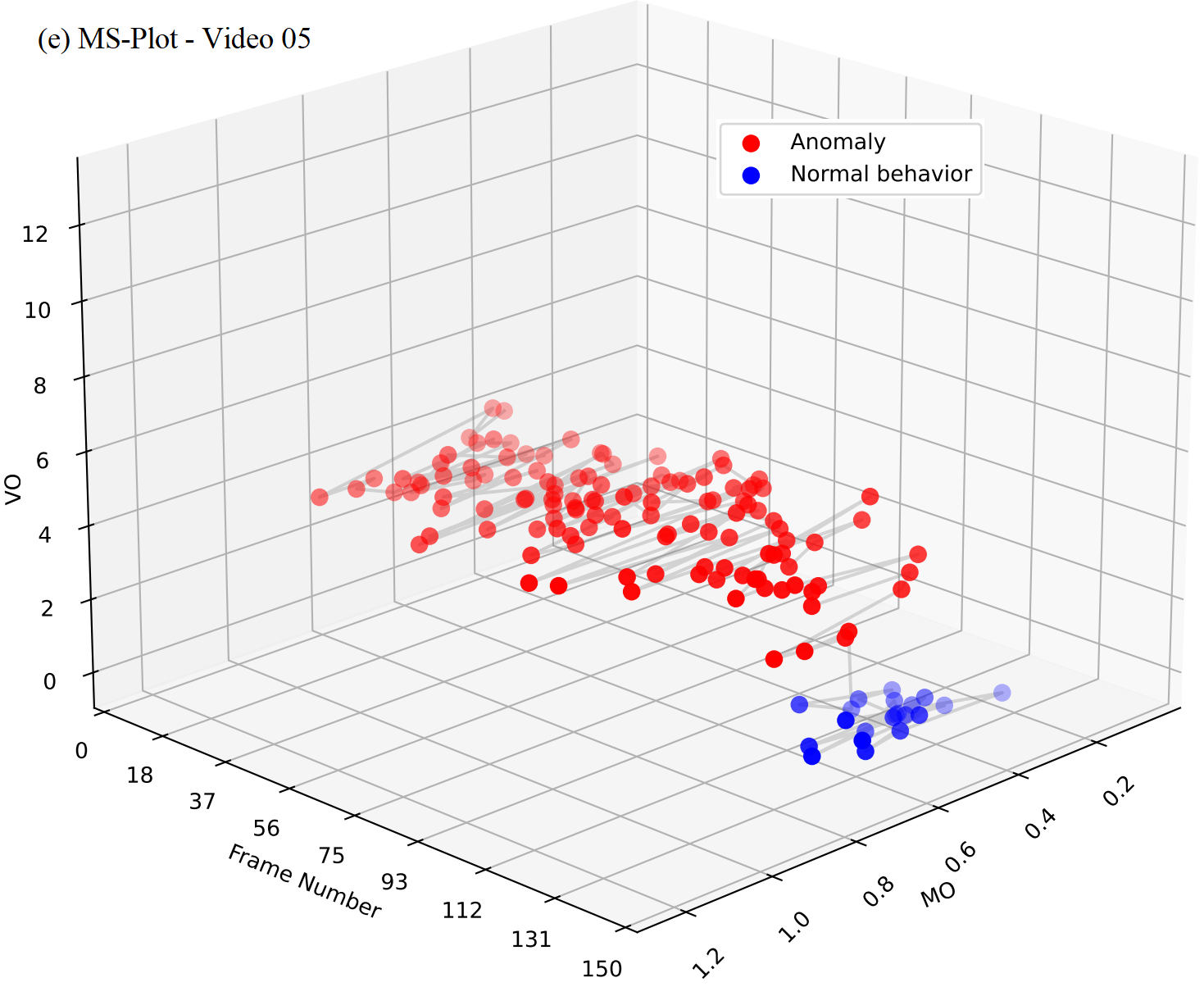}
    \includegraphics[width=6cm]{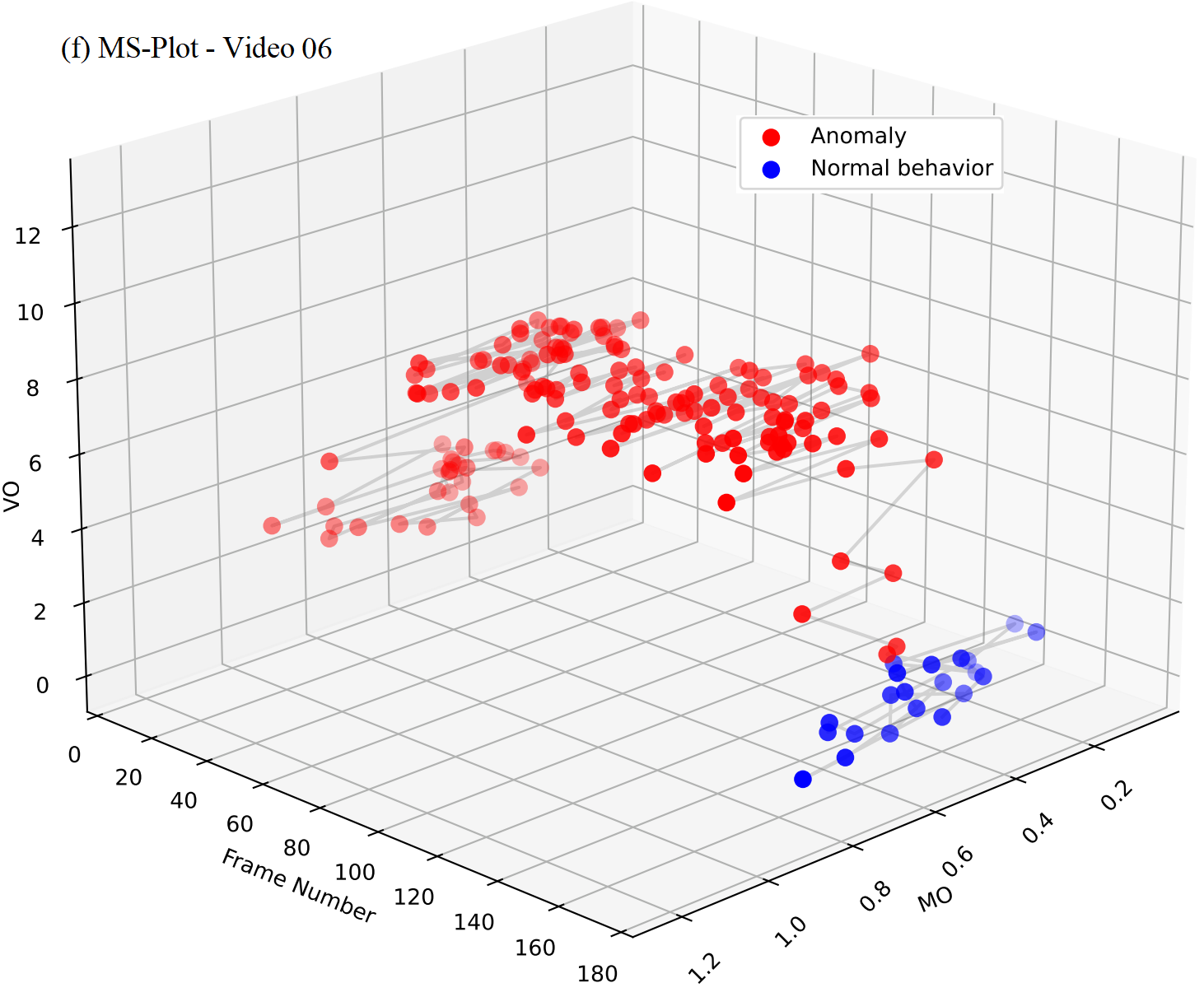}
    \includegraphics[width=6cm]{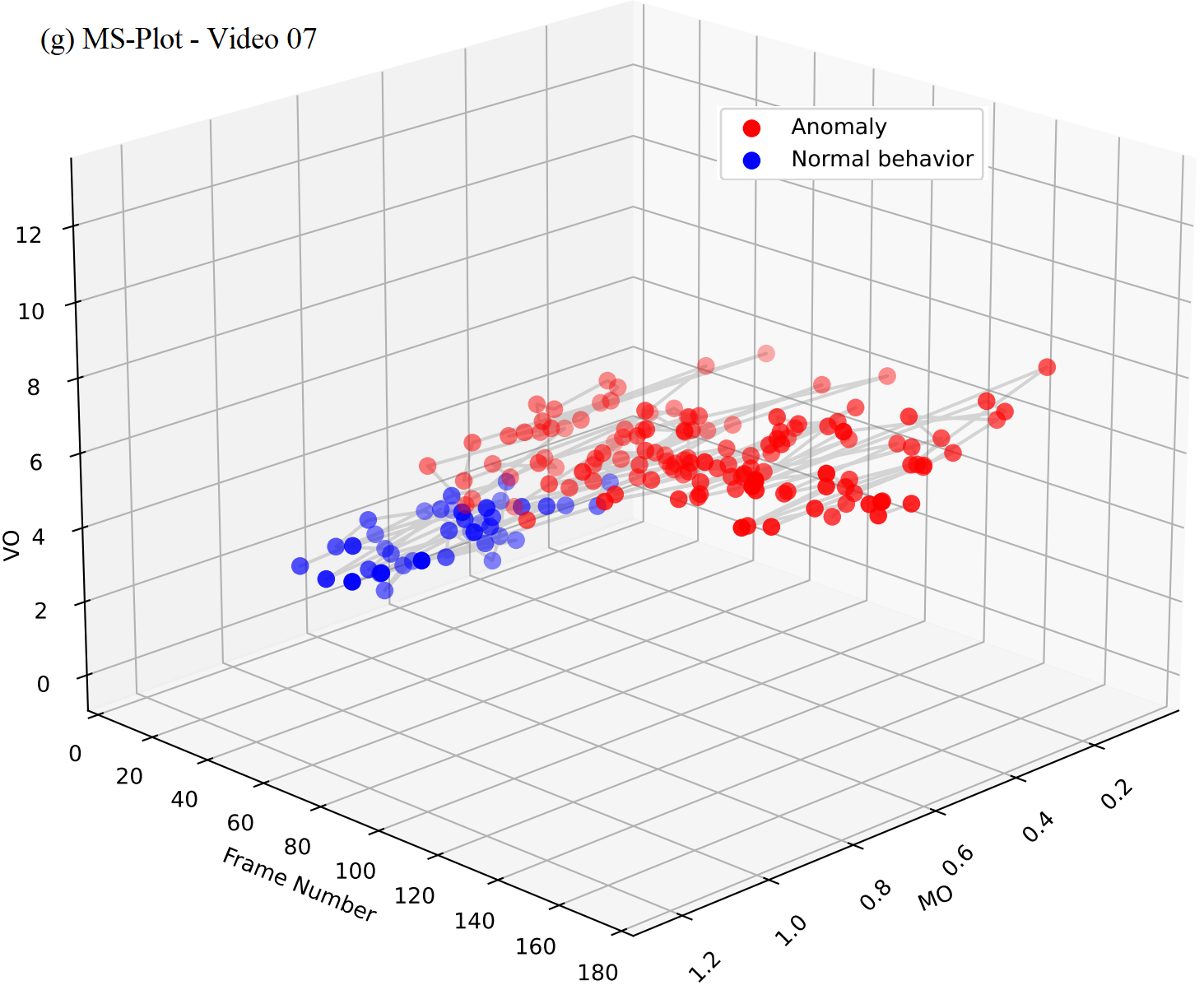}
     \includegraphics[width=6cm]{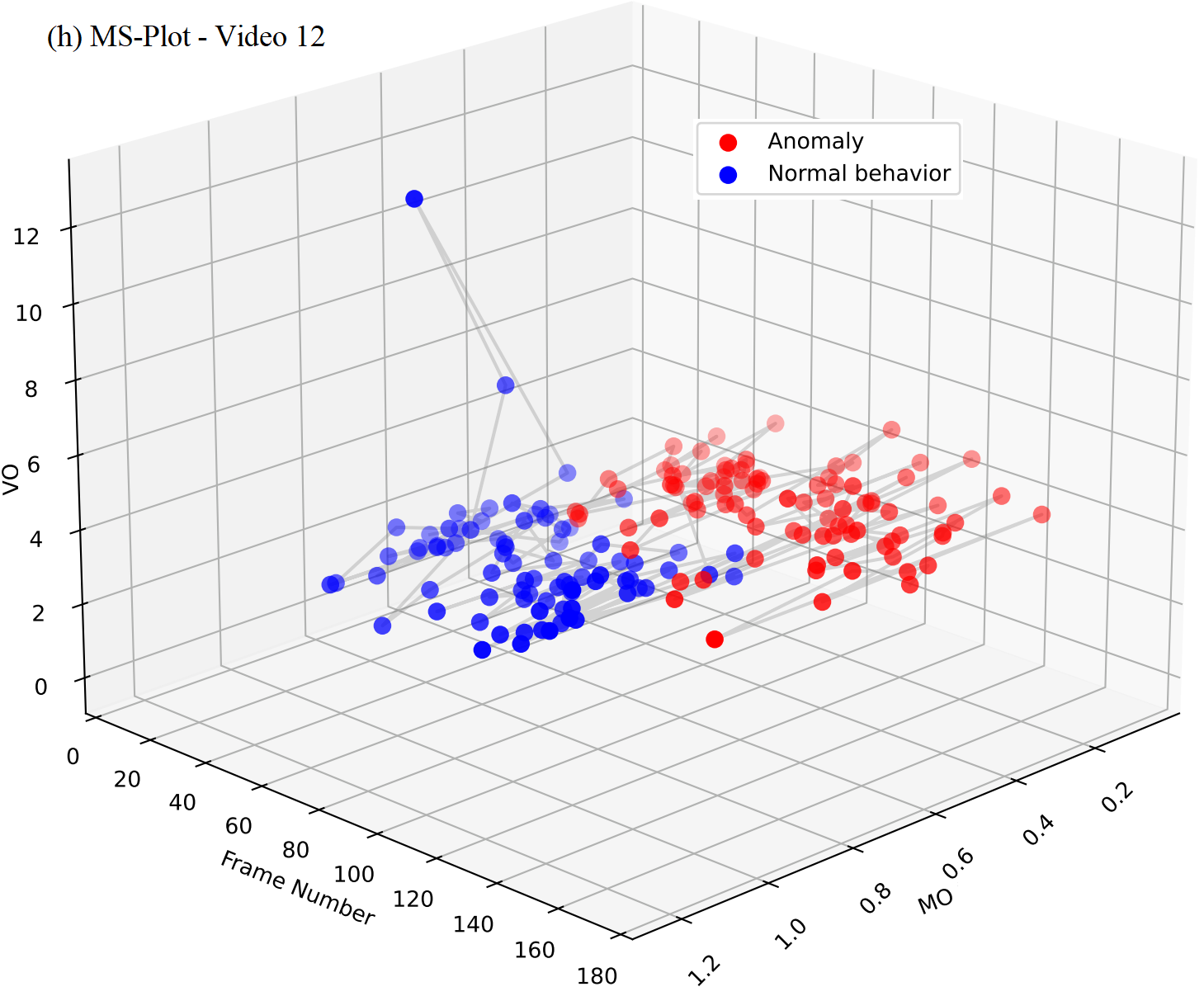}
  \caption{3D MS-Plot representation for videos in the UCSD Ped2 testing set. }
    \label{fig:enter-label}
\end{figure}

This study utilizes the UCSD Ped2 testing set to demonstrate the MS-Plot's ability to visually represent frame-level anomaly detection, as illustrated in Figure~\ref{fig:enter-label}. The 3D MS-Plot, with (\( |MO| \)) on the x-axis, (\( VO \)) on the y-axis, and frame numbers on the z-axis, highlights anomalies (red dots) and normal frames (blue dots) using residuals from the MAMA-based autoencoder. By excluding videos with solely anomalous frames (Table~\ref{tab:anomalous_frames}), the analysis highlights the MS-Plot’s ability to distinguish anomalies from normal behavior in mixed-event scenarios and track their progression over time.

Figure~\ref{fig:enter-label} illustrates the MS-Plot's capability to distinguish normal frames (blue points) from anomalies (red points) using magnitude outlyingness (\( |MO| \)) and shape outlyingness (\( VO \)). Normal frames cluster at lower \( |MO| \) and \( VO \) values, while anomalies show higher values, often forming distinct clusters. Elevated \( VO \) values highlight unusual shape variations, such as those caused by cyclists or skaters. The z-axis, representing frame numbers, tracks the temporal progression of anomalies, effectively visualizing their persistence and separation from normal behavior.

\newpage
\subsection{MS-Plot results based on CUHK Avenue data}
We extend the evaluation of the two investigated models, the AE-MS-Plot and MAMA-MS-Plot approaches, to the CUHK Avenue dataset to further analyze their performance. The CUHK Avenue dataset is particularly challenging due to its diverse set of anomalous behaviors, such as running, loitering, and abandoning objects, captured in a dynamic and semi-structured environment. By examining the performance of the MS-Plot-based methods on individual videos, the analysis provides insights into the robustness and sensitivity of these approaches to varying anomaly patterns and complexities within the dataset.  Following the same protocol as the UCSD Ped2 dataset, both models are trained exclusively on anomaly-free data to learn typical behavioral patterns and subsequently applied to the test set for anomaly detection. This methodology ensures that the models generalize effectively from normal data without prior exposure to anomalous events. The results from univariate functional detectors on the CUHK dataset were excluded due to their poor performance. The performance of the AE-MS-Plot and MAMA-MS-Plot models on the CUHK Avenue dataset is detailed in Tables~\ref{AECUHK} and~\ref{CUHK-MAMA}.  From Tables~\ref{AECUHK}, the AE-MS-Plot performs well on simpler anomalies (e.g., Videos 1, 4, and 12 with AUC > 97\%) but struggles with complex or predominantly anomalous frames (e.g., Videos 3, 10, and 19 with AUC of 50\%), highlighting its limitations in handling diverse scenarios. The unsatisfactory AE-based results may stem from dataset imbalances, particularly in CUHK Avenue, where the dominance of normal frames amplifies the impact of false negatives. Moreover, frames containing partial abnormal events, such as a small portion of a bike or an object at the start of a video, further complicate detection, revealing the limitations of the AE-based approach.
\medskip

\begin{table}[h!]
\centering
\caption{Performance of AE-MS-Plot on CUHK Test Videos.}\label{AECUHK}
\begin{tabular}{c|c|c|c|c|c|c}
\hline
Video & TPR  & FPR  & Accuracy & Precision  & F1-score  & AUC \\
\hline
1 & 94.90 & 0.00 & 98.40 & 100.00 & 97.38 & 97.70 \\
2 & 40.29 & 10.01 & 84.27 & 34.36 & 37.09 & 65.40 \\
3 & 100.00 & 100.00 & 9.35 & 9.35 & 17.10 & 50.00 \\
4 & 94.62 & 0.00 & 99.47 & 100.00 & 97.24 & 97.50 \\
5 & 92.14 & 0.00 & 97.51 & 100.00 & 95.91 & 96.30 \\
6 & 91.92 & 0.00 & 97.27 & 100.00 & 95.79 & 96.20 \\
7 & 67.62 & 0.00 & 94.35 & 100.00 & 80.68 & 84.00 \\
8 & 100.00 & 0.00 & 100.00 & 100.00 & 100.00 & 100.00 \\
9 & 15.92 & 0.00 & 76.11 & 100.00 & 27.46 & 58.20 \\
10 & 100.00 & 100.00 & 19.21 & 19.21 & 32.23 & 50.00 \\
11 & 100.00 & 0.00 & 100.00 & 100.00 & 100.00 & 100.00 \\
12 & 100.00 & 0.00 & 100.00 & 100.00 & 100.00 & 100.00 \\
13 & 92.59 & 0.00 & 98.90 & 100.00 & 96.15 & 96.50 \\
14 & 72.60 & 0.00 & 96.03 & 100.00 & 84.13 & 86.50 \\
15 & 84.44 & 20.04 & 80.36 & 29.46 & 43.68 & 82.40 \\
16 & 91.92 & 12.07 & 88.47 & 54.17 & 68.16 & 90.10 \\
17 & 79.61 & 0.00 & 82.74 & 100.00 & 88.65 & 90.00 \\
18 & 83.02 & 0.00 & 84.54 & 100.00 & 90.72 & 91.70 \\
19 & 100.00 & 100.00 & 53.88 & 53.88 & 70.03 & 50.00 \\
20 & 44.06 & 0.00 & 58.15 & 100.00 & 61.17 & 72.20 \\
21 & 64.15 & 10.00 & 71.23 & 94.44 & 76.40 & 77.30 \\
\hline
\end{tabular}
\end{table}

\medskip
The MAMA-MS-Plot performs well on the CUHK Avenue dataset, achieving high AUC values above 95\% in videos like 1, 4, 8, 10, and 13 (Table~\ref{CUHK-MAMA}). However, it faces challenges in videos with complex anomalies, such as 9 and 18, where AUC scores drop but remain above 88\%. These results indicate the method's overall reliability while highlighting areas for potential improvement in detecting anomalies in highly skewed scenarios.
\begin{table}[h!]
\centering
\caption{Performance of MAMA-MS-Plot on CUHK Test Videos.}
\label{CUHK-MAMA}
\begin{tabular}{c|c|c|c|c|c|c}
\hline
Video & TPR  & FPR & Accuracy & Precision & F1-score & AUC \\
\hline
1 & 84.92 & 3.15 & 93.11 & 92.51 & 88.55 & 91.21 \\
2 & 84.89 & 3.18 & 95.45 & 77.63 & 81.10 & 91.18 \\
3 & 95.35 & 3.36 & 96.52 & 74.55 & 83.67 & 96.32 \\
4 & 93.55 & 1.65 & 97.88 & 86.14 & 89.69 & 96.27 \\
5 & 82.70 & 1.02 & 93.82 & 97.41 & 89.46 & 91.16 \\
6 & 89.15 & 1.18 & 95.55 & 97.47 & 93.12 & 94.30 \\
7 & 80.95 & 1.01 & 95.85 & 94.44 & 87.18 & 90.29 \\
8 & 90.00 & 0.00 & 96.97 & 100.00 & 94.74 & 95.32 \\
9 & 76.88 & 0.72 & 92.92 & 97.71 & 86.05 & 88.40 \\
10 & 93.79 & 0.30 & 98.57 & 98.69 & 96.18 & 97.07 \\
11 & 81.42 & 0.00 & 92.75 & 100.00 & 89.76 & 91.03 \\
12 & 79.92 & 0.79 & 95.43 & 96.14 & 87.28 & 89.89 \\
13 & 95.06 & 0.00 & 99.27 & 100.00 & 97.47 & 97.85 \\
14 & 90.41 & 2.09 & 96.83 & 88.00 & 89.19 & 94.48 \\
15 & 96.67 & 2.75 & 97.19 & 77.68 & 86.14 & 97.28 \\
16 & 91.92 & 1.25 & 97.83 & 91.92 & 91.92 & 95.65 \\
17 & 77.93 & 0.00 & 81.32 & 100.00 & 87.60 & 89.29 \\
18 & 78.49 & 0.00 & 80.41 & 100.00 & 87.95 & 89.57 \\
19 & 80.30 & 1.77 & 88.57 & 98.15 & 88.33 & 89.59 \\
20 & 78.71 & 0.00 & 84.07 & 100.00 & 88.09 & 89.68 \\
21 & 79.25 & 0.00 & 84.93 & 100.00 & 88.42 & 89.94 \\
\hline
\end{tabular}
\end{table}

The overall AUC results show an improvement with the MAMA-MS-Plot (91.43\%) over the AE-MS-Plot (81.40\%), demonstrating enhanced anomaly detection performance through its advanced residual analysis combined with the MS-Plot framework.

\newpage
\subsection{Comparison with SOTA methods}
Table~\ref{tab:comparison_state_of_art_ucsdped2} provides an overall comparison of AUC scores for the proposed MS-Plot-based approaches and state-of-the-art (SOTA) methods on the UCSD Ped2 dataset. The MAMA-MS-Plot achieves an impressive AUC of 98.74\%, while the AE-MS-Plot records 88.30\%, showcasing the overall effectiveness of the MS-Plot framework in accurately identifying video anomalies. Specifically, compared to other detection methods, the MAMA-MS-Plot demonstrated the highest performance, achieving the best AUC score among the evaluated approaches.
\begin{table}[h!]
\centering
\caption{Comparison of AUC (\%) with SOTA methods on the UCSD Ped2 dataset.}
\label{tab:comparison_state_of_art_ucsdped2}
\begin{tabular}{lr}
\toprule
\textbf{Method} & \textbf{AUC (\%)} \\
\midrule
MPPCA~\cite{kim2009observe} & 77.00 \\
Motion-appearance model~\cite{zhang2016combining} & 90.00 \\
Spatial Temporal CNN~\cite{zhou2016spatial} & 86.00 \\
Conv-AE~\cite{hasan2016learning} & 90.00 \\
AMDN~\cite{xu2015learning} & 90.80 \\
GMFC-VAE~\cite{fan2020video} & 92.20 \\
StackRNN~\cite{luo2017revisit} & 92.20\\
MGFC-AAE~\cite{li2019video} & 91.60 \\
ST-CaAE~\cite{li2020spatial} & 92.90 \\
MemAE~\cite{gong2019memorizing}  &  94.10 \\
DAW~\cite{wang2018detecting} &   96.40 \\
MPN~\cite{lv2021learning} & 96.90\\
ConvLSTM~\cite{luo2017remembering} & 88.10\\
spatiotemporal AE (STAE)~\cite{zhao2017spatio} & 91.20\\
Conv2D~\cite{hasan2016learning} & 90.00\\
TMAE~\cite{hu2022detecting} & 94.10\\
MAMC~\cite{ning2024memory} & 96.70\\
Cascade Reconstruction~\cite{zhong2022cascade} & 97.70\\
MAMA~\cite{hong2024making} & 98.20 \\
\hline
AE-MS-Plot & 88.30\\
MAMA-MS-Plot & 98.74 \\
\bottomrule
\end{tabular}
\end{table}

The MAMA-MS-Plot method demonstrates improved performance over the original MAMA-based approach, achieving an AUC of 98.74\% compared to 98.20\%. This improvement highlights the benefits of using the MS-Plot framework for anomaly detection, particularly in its ability to enhance the interpretability and precision of detection outcomes. In the original MAMA model, anomaly detection is based on empirically determined thresholds primarily based on abnormal changes in the magnitude of residuals. These thresholds, while functional, lack a systematic foundation, which can lead to inconsistencies when addressing diverse anomaly patterns or capturing more nuanced deviations in behavior. The integration of the MS-Plot framework addresses this limitation by treating the residuals of the MAMA model as multivariate functional data. The MS plot evaluates outlyingness (amplitude deviations) and shape (pattern deviations), providing a multidimensional analysis of residual patterns. This capability is particularly beneficial in video anomaly detection, where anomalies often exhibit subtle or complex deviations that may not be effectively captured by simpler thresholding techniques. Using the MS-Plot framework, the MAMA-MS-Plot approach offers a statistically principled method to analyze residuals, enhancing the consistency and reliability of anomaly detection. The framework’s capacity to capture both amplitude and shape deviations allows for a more comprehensive understanding of anomalies, ultimately improving detection performance in complex and dynamic video environments.

\medskip 
We compare the proposed MS-Plot-based methods with several state-of-the-art anomaly detection techniques in the CUHK Avenue dataset, as detailed in Table~\ref{tab:comparison_state_of_art_cuhk}. The MAMA-MS-Plot approach achieves an AUC of 91.43\%, demonstrating its strong competitiveness among SOTA methods. In particular, the previous highest reported AUC was 91.30\%, achieved by SD-MAE~\cite{ristea2024self}. The MAMA-MS-Plot exceeds this by a small but meaningful margin of 0.13\%, setting a new benchmark for anomaly detection performance in this data set.

\begin{table}[h!]
\centering
\caption{Comparison of AUC (\%) with SOTA methods on the CUHK Avenue dataset.}
\label{tab:comparison_state_of_art_cuhk}
\begin{tabular}{lr}
\toprule
\textbf{Method} & \textbf{AUC (\%)} \\
\midrule
Unmasking~\cite{tudor2017unmasking} & 80.60 \\
StackRNN~\cite{luo2017revisit} & 81.70\\
MemAE~\cite{gong2019memorizing} & 83.3 \\
FastAno~\cite{park2022fastano} & 85.30 \\

STemGAN~\cite{singh2023stemgan} & 86.00 \\	
EVAL ~\cite{singh2023eval}& 86.02\\
MESDnet~\cite{fang2020multi} & 86.3 \\
Any-Shot Sequential ~\cite{doshi2020any}& 86.40\\
AMMC-Net~\cite{cai2021appearance} & 86.60 \\
ASTNet ~\cite{le2023attention}& 86.70 \\
Context Pre~\cite{li2022context} & 87.10\\
Learning not to reconstruct ~\cite{astrid2021learning}& 87.10 \\
Siamese Net~\cite{ramachandra2020learning} & 87.20 \\
Object-centric AE~\cite{ionescu2019object} & 87.40 \\
AKD-VAD~\cite{croitoru2022lightning} & 88.30 \\
AnomalyRuler~\cite{yang2025follow} &89.70 \\
Two-stream~\cite{cao2024context} &90.80 \\	
MAMA~\cite{hong2024making} & 91.20 \\
SD-MAE~\cite{ristea2024self} & 91.30 \\
\hline
AE-MS-Plot & 81.40\\
MAMA-MS-Plot & 91.43 \\
\bottomrule
\end{tabular}
\end{table}

\medskip
This improvement, while incremental, underscores the value of integrating the MS-Plot framework, which provides a systematic and statistically principled approach to anomaly detection. By treating residuals as multivariate functional data, the MAMA-MS-Plot captures both magnitude and shape deviations, enabling more precise differentiation between normal and anomalous frames. In contrast, traditional methods often rely on empirically determined thresholds or univariate analyses, which may miss subtle or multidimensional patterns indicative of anomalies. The MS-Plot's ability to enhance the performance of already strong models like MAMA highlights its potential as a generalizable and effective addition to video anomaly detection pipelines. This advantage is particularly significant given the challenging nature of the CUHK Avenue dataset, which features diverse and complex anomalous behaviors.

\section{Conclusion}\label{sec5}
This study demonstrates the efficacy of the Magnitude-Shape (MS) Plot framework for video anomaly detection, combining statistical functional data analysis with reconstruction-based models. By treating reconstruction errors as multivariate functional data, the MS-Plot captures both magnitude and shape deviations, enabling accurate identification of anomalies in complex, crowded video scenes. The integration of autoencoders with the MS-Plot enhances anomaly detection by leveraging their capacity to model normal behavior and identify deviations with statistical rigor. Experimental results on benchmark datasets, including UCSD Ped2 and CUHK Avenue, highlight the advantages of the proposed framework over traditional univariate functional detectors and several state-of-the-art methods. Specifically, the MAMA-MS-Plot approach achieves consistently high AUC scores, showcasing its capability to generalize across diverse scenarios and detect both subtle and pronounced anomalies effectively.

\medskip 
The findings emphasize the potential of MS-Plot-based frameworks to address the challenges of video anomaly detection, such as the variability of anomalies and limited labeled data.  An important direction for future work is to focus on developing unsupervised and robust statistical methods that do not rely on anomaly-free training data. Such methods would further enhance the practicality of video anomaly detection by eliminating the dependence on curated normal datasets and addressing the inherent challenges of real-world applications, such as data scarcity and variability in anomaly types.

\bibliographystyle{unsrtnat}
\bibliography{Manuscript}

\begin{thebibliography}{89}
\providecommand{\natexlab}[1]{#1}
\providecommand{\url}[1]{\texttt{#1}}
\expandafter\ifx\csname urlstyle\endcsname\relax
  \providecommand{\doi}[1]{doi: #1}\else
  \providecommand{\doi}{doi: \begingroup \urlstyle{rm}\Url}\fi

\bibitem[Santhosh et~al.(2020)Santhosh, Dogra, and Roy]{santhosh2020anomaly}
Kelathodi~Kumaran Santhosh, Debi~Prosad Dogra, and Partha~Pratim Roy.
\newblock Anomaly detection in road traffic using visual surveillance: A survey.
\newblock \emph{ACM Computing Surveys (CSUR)}, 53\penalty0 (6):\penalty0 1--26, 2020.

\bibitem[Tripathi et~al.(2018)Tripathi, Jalal, and Agrawal]{tripathi2018suspicious}
Rajesh~Kumar Tripathi, Anand~Singh Jalal, and Subhash~Chand Agrawal.
\newblock Suspicious human activity recognition: a review.
\newblock \emph{Artificial Intelligence Review}, 50:\penalty0 283--339, 2018.

\bibitem[Chandola et~al.(2009)Chandola, Banerjee, and Kumar]{chandola2009anomaly}
Varun Chandola, Arindam Banerjee, and Vipin Kumar.
\newblock Anomaly detection: A survey.
\newblock \emph{ACM computing surveys (CSUR)}, 41\penalty0 (3):\penalty0 1--58, 2009.

\bibitem[Nayak et~al.(2021)Nayak, Pati, and Das]{nayak2021comprehensive}
Rashmiranjan Nayak, Umesh~Chandra Pati, and Santos~Kumar Das.
\newblock A comprehensive review on deep learning-based methods for video anomaly detection.
\newblock \emph{Image and Vision Computing}, 106:\penalty0 104078, 2021.

\bibitem[Mu et~al.(2021)Mu, Sun, Yuan, and Wang]{mu2021abnormal}
Huiyu Mu, Ruizhi Sun, Gang Yuan, and Yun Wang.
\newblock Abnormal human behavior detection in videos: A review.
\newblock \emph{Information Technology and Control}, 50\penalty0 (3):\penalty0 522--545, 2021.

\bibitem[Duong et~al.(2023)Duong, Le, and Hoang]{duong2023deep}
Huu-Thanh Duong, Viet-Tuan Le, and Vinh~Truong Hoang.
\newblock Deep learning-based anomaly detection in video surveillance: A survey.
\newblock \emph{Sensors}, 23\penalty0 (11):\penalty0 5024, 2023.

\bibitem[Zhu et~al.(2012)Zhu, Nayak, and Roy-Chowdhury]{zhu2012context}
Yingying Zhu, Nandita~M Nayak, and Amit~K Roy-Chowdhury.
\newblock Context-aware activity recognition and anomaly detection in video.
\newblock \emph{IEEE Journal of Selected Topics in Signal Processing}, 7\penalty0 (1):\penalty0 91--101, 2012.

\bibitem[Pawar and Attar(2019)]{pawar2019deep}
Karishma Pawar and Vahida Attar.
\newblock Deep learning approaches for video-based anomalous activity detection.
\newblock \emph{World Wide Web}, 22\penalty0 (2):\penalty0 571--601, 2019.

\bibitem[Morris and Trivedi(2008)]{morris2008survey}
Brendan~Tran Morris and Mohan~Manubhai Trivedi.
\newblock A survey of vision-based trajectory learning and analysis for surveillance.
\newblock \emph{IEEE transactions on circuits and systems for video technology}, 18\penalty0 (8):\penalty0 1114--1127, 2008.

\bibitem[Kim and Grauman(2009)]{kim2009observe}
Jaechul Kim and Kristen Grauman.
\newblock Observe locally, infer globally: a space-time mrf for detecting abnormal activities with incremental updates.
\newblock In \emph{2009 IEEE conference on computer vision and pattern recognition}, pages 2921--2928. IEEE, 2009.

\bibitem[Mehran et~al.(2009)Mehran, Oyama, and Shah]{mehran2009abnormal}
Ramin Mehran, Alexis Oyama, and Mubarak Shah.
\newblock Abnormal crowd behavior detection using social force model.
\newblock In \emph{2009 IEEE conference on computer vision and pattern recognition}, pages 935--942. IEEE, 2009.

\bibitem[Basharat et~al.(2008)Basharat, Gritai, and Shah]{basharat2008learning}
Arslan Basharat, Alexei Gritai, and Mubarak Shah.
\newblock Learning object motion patterns for anomaly detection and improved object detection.
\newblock In \emph{2008 IEEE conference on computer vision and pattern recognition}, pages 1--8. IEEE, 2008.

\bibitem[Piciarelli et~al.(2008)Piciarelli, Micheloni, and Foresti]{piciarelli2008trajectory}
Claudio Piciarelli, Christian Micheloni, and Gian~Luca Foresti.
\newblock Trajectory-based anomalous event detection.
\newblock \emph{IEEE Transactions on Circuits and Systems for video Technology}, 18\penalty0 (11):\penalty0 1544--1554, 2008.

\bibitem[Co{\c{s}}ar et~al.(2016)Co{\c{s}}ar, Donatiello, Bogorny, Garate, Alvares, and Br{\'e}mond]{cocsar2016toward}
Serhan Co{\c{s}}ar, Giuseppe Donatiello, Vania Bogorny, Carolina Garate, Luis~Otavio Alvares, and Fran{\c{c}}ois Br{\'e}mond.
\newblock Toward abnormal trajectory and event detection in video surveillance.
\newblock \emph{IEEE Transactions on Circuits and Systems for Video Technology}, 27\penalty0 (3):\penalty0 683--695, 2016.

\bibitem[Ma et~al.(2014)Ma, Wang, Xue, Zhou, Ji, and Li]{ma2014depth}
Xin Ma, Haibo Wang, Bingxia Xue, Mingang Zhou, Bing Ji, and Yibin Li.
\newblock Depth-based human fall detection via shape features and improved extreme learning machine.
\newblock \emph{IEEE journal of biomedical and health informatics}, 18\penalty0 (6):\penalty0 1915--1922, 2014.

\bibitem[Li et~al.(2013)Li, Mahadevan, and Vasconcelos]{li2013anomaly}
Weixin Li, Vijay Mahadevan, and Nuno Vasconcelos.
\newblock Anomaly detection and localization in crowded scenes.
\newblock \emph{IEEE transactions on pattern analysis and machine intelligence}, 36\penalty0 (1):\penalty0 18--32, 2013.

\bibitem[Lu et~al.(2013)Lu, Shi, and Jia]{lu2013abnormal}
Cewu Lu, Jianping Shi, and Jiaya Jia.
\newblock Abnormal event detection at 150 fps in matlab.
\newblock In \emph{Proceedings of the IEEE international conference on computer vision}, pages 2720--2727, 2013.

\bibitem[Dalal and Triggs(2005)]{dalal2005histograms}
Navneet Dalal and Bill Triggs.
\newblock Histograms of oriented gradients for human detection.
\newblock In \emph{2005 IEEE computer society conference on computer vision and pattern recognition (CVPR'05)}, volume~1, pages 886--893. Ieee, 2005.

\bibitem[Kratz and Nishino(2009)]{kratz2009anomaly}
Louis Kratz and Ko~Nishino.
\newblock Anomaly detection in extremely crowded scenes using spatio-temporal motion pattern models.
\newblock In \emph{2009 IEEE conference on computer vision and pattern recognition}, pages 1446--1453. IEEE, 2009.

\bibitem[Dalal et~al.(2006)Dalal, Triggs, and Schmid]{dalal2006human}
Navneet Dalal, Bill Triggs, and Cordelia Schmid.
\newblock Human detection using oriented histograms of flow and appearance.
\newblock In \emph{Computer Vision--ECCV 2006: 9th European Conference on Computer Vision, Graz, Austria, May 7-13, 2006. Proceedings, Part II 9}, pages 428--441. Springer, 2006.

\bibitem[Roy and Om(2018)]{roy2018suspicious}
Produte~Kumar Roy and Hari Om.
\newblock Suspicious and violent activity detection of humans using hog features and svm classifier in surveillance videos.
\newblock \emph{Advances in Soft Computing and Machine Learning in Image Processing}, pages 277--294, 2018.

\bibitem[Ahad et~al.(2018)Ahad, Tan, Kim, and Ishikawa]{ahad2018activity}
Md~Atiqur~Rahman Ahad, Joo~Kooi Tan, Hyoungseop Kim, and Seiji Ishikawa.
\newblock Activity representation by surf-based templates.
\newblock \emph{Computer Methods in Biomechanics and Biomedical Engineering: Imaging \& Visualization}, 6\penalty0 (5):\penalty0 573--583, 2018.

\bibitem[Cheng et~al.(2015)Cheng, Chen, and Fang]{cheng2015video}
Kai-Wen Cheng, Yie-Tarng Chen, and Wen-Hsien Fang.
\newblock Video anomaly detection and localization using hierarchical feature representation and gaussian process regression.
\newblock In \emph{Proceedings of the IEEE Conference on Computer Vision and Pattern Recognition}, pages 2909--2917, 2015.

\bibitem[Sabokrou et~al.(2018)Sabokrou, Fayyaz, Fathy, Moayed, and Klette]{sabokrou2018deep}
Mohammad Sabokrou, Mohsen Fayyaz, Mahmood Fathy, Zahra Moayed, and Reinhard Klette.
\newblock Deep-anomaly: Fully convolutional neural network for fast anomaly detection in crowded scenes.
\newblock \emph{Computer Vision and Image Understanding}, 172:\penalty0 88--97, 2018.

\bibitem[Mansour et~al.(2021)Mansour, Escorcia-Gutierrez, Gamarra, Villanueva, and Leal]{mansour2021intelligent}
Romany~F Mansour, Jos{\'e} Escorcia-Gutierrez, Margarita Gamarra, Jair~A Villanueva, and Nallig Leal.
\newblock Intelligent video anomaly detection and classification using faster rcnn with deep reinforcement learning model.
\newblock \emph{Image and Vision Computing}, 112:\penalty0 104229, 2021.

\bibitem[Sultani et~al.(2018)Sultani, Chen, and Shah]{sultani2018real}
Waqas Sultani, Chen Chen, and Mubarak Shah.
\newblock Real-world anomaly detection in surveillance videos.
\newblock In \emph{Proceedings of the IEEE conference on computer vision and pattern recognition}, pages 6479--6488, 2018.

\bibitem[Nawaratne et~al.(2019)Nawaratne, Alahakoon, De~Silva, and Yu]{nawaratne2019spatiotemporal}
Rashmika Nawaratne, Damminda Alahakoon, Daswin De~Silva, and Xinghuo Yu.
\newblock Spatiotemporal anomaly detection using deep learning for real-time video surveillance.
\newblock \emph{IEEE Transactions on Industrial Informatics}, 16\penalty0 (1):\penalty0 393--402, 2019.

\bibitem[Ullah et~al.(2021)Ullah, Ullah, Hussain, Khan, and Baik]{ullah2021efficient}
Waseem Ullah, Amin Ullah, Tanveer Hussain, Zulfiqar~Ahmad Khan, and Sung~Wook Baik.
\newblock An efficient anomaly recognition framework using an attention residual lstm in surveillance videos.
\newblock \emph{Sensors}, 21\penalty0 (8):\penalty0 2811, 2021.

\bibitem[Luo et~al.(2021)Luo, Liu, Lian, and Gao]{luo2021future}
Weixin Luo, Wen Liu, Dongze Lian, and Shenghua Gao.
\newblock Future frame prediction network for video anomaly detection.
\newblock \emph{IEEE transactions on pattern analysis and machine intelligence}, 44\penalty0 (11):\penalty0 7505--7520, 2021.

\bibitem[Micorek et~al.(2024)Micorek, Possegger, Narnhofer, Bischof, and Kozinski]{micorek2024mulde}
Jakub Micorek, Horst Possegger, Dominik Narnhofer, Horst Bischof, and Mateusz Kozinski.
\newblock Mulde: Multiscale log-density estimation via denoising score matching for video anomaly detection.
\newblock In \emph{Proceedings of the IEEE/CVF Conference on Computer Vision and Pattern Recognition}, pages 18868--18877, 2024.

\bibitem[Chen et~al.(2022)Chen, Duan, Kang, and Qiu]{chen2022supervised}
Zhi Chen, Jiang Duan, Li~Kang, and Guoping Qiu.
\newblock Supervised anomaly detection via conditional generative adversarial network and ensemble active learning.
\newblock \emph{IEEE Transactions on Pattern Analysis and Machine Intelligence}, 45\penalty0 (6):\penalty0 7781--7798, 2022.

\bibitem[Wu et~al.(2024)Wu, Pan, Yan, Pang, Wang, and Zhang]{wu2024deep}
Peng Wu, Chengyu Pan, Yuting Yan, Guansong Pang, Peng Wang, and Yanning Zhang.
\newblock Deep learning for video anomaly detection: A review.
\newblock \emph{arXiv preprint arXiv:2409.05383}, 2024.

\bibitem[Zhou et~al.(2016)Zhou, Shen, Zeng, Fang, Wei, and Zhang]{zhou2016spatial}
Shifu Zhou, Wei Shen, Dan Zeng, Mei Fang, Yuanwang Wei, and Zhijiang Zhang.
\newblock Spatial--temporal convolutional neural networks for anomaly detection and localization in crowded scenes.
\newblock \emph{Signal Processing: Image Communication}, 47:\penalty0 358--368, 2016.

\bibitem[Hong et~al.(2024)Hong, Ahn, Jo, and Park]{hong2024making}
Seungkyun Hong, Sunghyun Ahn, Youngwan Jo, and Sanghyun Park.
\newblock Making anomalies more anomalous: Video anomaly detection using a novel generator and destroyer.
\newblock \emph{IEEE Access}, 2024.

\bibitem[Dilek and Dener(2024)]{dilek2024enhancement}
Esma Dilek and Murat Dener.
\newblock Enhancement of video anomaly detection performance using transfer learning and fine-tuning.
\newblock \emph{IEEE Access}, 2024.

\bibitem[Chong and Tay(2017)]{chong2017abnormal}
Yong~Shean Chong and Yong~Haur Tay.
\newblock Abnormal event detection in videos using spatiotemporal autoencoder.
\newblock In \emph{Advances in Neural Networks-ISNN 2017: 14th International Symposium, ISNN 2017, Sapporo, Hakodate, and Muroran, Hokkaido, Japan, June 21--26, 2017, Proceedings, Part II 14}, pages 189--196. Springer, 2017.

\bibitem[Gong et~al.(2019)Gong, Liu, Le, Saha, Mansour, Venkatesh, and Hengel]{gong2019memorizing}
Dong Gong, Lingqiao Liu, Vuong Le, Budhaditya Saha, Moussa~Reda Mansour, Svetha Venkatesh, and Anton van~den Hengel.
\newblock Memorizing normality to detect anomaly: Memory-augmented deep autoencoder for unsupervised anomaly detection.
\newblock In \emph{Proceedings of the IEEE/CVF international conference on computer vision}, pages 1705--1714, 2019.

\bibitem[Le and Kim(2023)]{le2023attention}
Viet-Tuan Le and Yong-Guk Kim.
\newblock Attention-based residual autoencoder for video anomaly detection.
\newblock \emph{Applied Intelligence}, 53\penalty0 (3):\penalty0 3240--3254, 2023.

\bibitem[Zhao et~al.(2017)Zhao, Deng, Shen, Liu, Lu, and Hua]{zhao2017spatio}
Yiru Zhao, Bing Deng, Chen Shen, Yao Liu, Hongtao Lu, and Xian-Sheng Hua.
\newblock Spatio-temporal autoencoder for video anomaly detection.
\newblock In \emph{Proceedings of the 25th ACM international conference on Multimedia}, pages 1933--1941, 2017.

\bibitem[Deepak et~al.(2021)Deepak, Chandrakala, and Mohan]{deepak2021residual}
K~Deepak, S~Chandrakala, and C~Krishna Mohan.
\newblock Residual spatiotemporal autoencoder for unsupervised video anomaly detection.
\newblock \emph{Signal, Image and Video Processing}, 15\penalty0 (1):\penalty0 215--222, 2021.

\bibitem[Luo et~al.(2019)Luo, Liu, Lian, Tang, Duan, Peng, and Gao]{luo2019video}
Weixin Luo, Wen Liu, Dongze Lian, Jinhui Tang, Lixin Duan, Xi~Peng, and Shenghua Gao.
\newblock Video anomaly detection with sparse coding inspired deep neural networks.
\newblock \emph{IEEE transactions on pattern analysis and machine intelligence}, 43\penalty0 (3):\penalty0 1070--1084, 2019.

\bibitem[Kumar and Khari(2023)]{kumar2023efficient}
Ankit Kumar and Manju Khari.
\newblock Efficient video anomaly detection using residual variational autoencoder.
\newblock In \emph{2023 International Conference on Communication System, Computing and IT Applications (CSCITA)}, pages 50--55. IEEE, 2023.

\bibitem[Aslam and Kolekar(2024)]{aslam2024demaae}
Nazia Aslam and Maheshkumar~H Kolekar.
\newblock Demaae: deep multiplicative attention-based autoencoder for identification of peculiarities in video sequences.
\newblock \emph{The Visual Computer}, 40\penalty0 (3):\penalty0 1729--1743, 2024.

\bibitem[Li and Chang(2019)]{li2019video}
Nanjun Li and Faliang Chang.
\newblock Video anomaly detection and localization via multivariate gaussian fully convolution adversarial autoencoder.
\newblock \emph{Neurocomputing}, 369:\penalty0 92--105, 2019.

\bibitem[Hu et~al.(2022{\natexlab{a}})Hu, Lian, Zhang, Gao, Jiang, and Chen]{hu2022video}
Xing Hu, Jing Lian, Dawei Zhang, Xiumin Gao, Linhua Jiang, and Wenmin Chen.
\newblock Video anomaly detection based on 3d convolutional auto-encoder.
\newblock \emph{Signal, Image and Video Processing}, 16\penalty0 (7):\penalty0 1885--1893, 2022{\natexlab{a}}.

\bibitem[Duman and Erdem(2019)]{duman2019anomaly}
Elvan Duman and Osman~Ayhan Erdem.
\newblock Anomaly detection in videos using optical flow and convolutional autoencoder.
\newblock \emph{IEEE Access}, 7:\penalty0 183914--183923, 2019.

\bibitem[Mishra and Jabin(2024)]{mishra2024anomaly}
Sarthak Mishra and Suraiya Jabin.
\newblock Anomaly detection in surveillance videos using deep autoencoder.
\newblock \emph{International Journal of Information Technology}, 16\penalty0 (2):\penalty0 1111--1122, 2024.

\bibitem[Dai and Genton(2018)]{dai2018multivariate}
Wenlin Dai and Marc~G Genton.
\newblock Multivariate functional data visualization and outlier detection.
\newblock \emph{Journal of Computational and Graphical Statistics}, 27\penalty0 (4):\penalty0 923--934, 2018.

\bibitem[Rousseeuw et~al.(2018)Rousseeuw, Raymaekers, and Hubert]{rousseeuw2018measure}
Peter~J Rousseeuw, Jakob Raymaekers, and Mia Hubert.
\newblock A measure of directional outlyingness with applications to image data and video.
\newblock \emph{Journal of Computational and Graphical Statistics}, 27\penalty0 (2):\penalty0 345--359, 2018.

\bibitem[Alem{\'a}n-G{\'o}mez et~al.(2022)Alem{\'a}n-G{\'o}mez, Arribas-Gil, Desco, El{\'\i}as, and Romo]{aleman2022depthgram}
Yasser Alem{\'a}n-G{\'o}mez, Ana Arribas-Gil, Manuel Desco, Antonio El{\'\i}as, and Juan Romo.
\newblock Depthgram: Visualizing outliers in high-dimensional functional data with application to fmri data exploration.
\newblock \emph{Statistics in Medicine}, 41\penalty0 (11):\penalty0 2005--2024, 2022.

\bibitem[Naji et~al.(2022)Naji, Setkov, Loesch, Gouiff{\`e}s, and Audigier]{naji2022spatio}
Yassine Naji, Aleksandr Setkov, Angelique Loesch, Mich{\`e}le Gouiff{\`e}s, and Romaric Audigier.
\newblock Spatio-temporal predictive tasks for abnormal event detection in videos.
\newblock In \emph{2022 18th IEEE International Conference on Advanced Video and Signal Based Surveillance (AVSS)}, pages 1--8. IEEE, 2022.

\bibitem[Yang et~al.(2025)Yang, Lee, Dariush, Cao, and Lo]{yang2025follow}
Yuchen Yang, Kwonjoon Lee, Behzad Dariush, Yinzhi Cao, and Shao-Yuan Lo.
\newblock Follow the rules: reasoning for video anomaly detection with large language models.
\newblock In \emph{European Conference on Computer Vision}, pages 304--322. Springer, 2025.

\bibitem[Singh et~al.(2023{\natexlab{a}})Singh, Saini, Sethi, Tiwari, Saurav, and Singh]{singh2023stemgan}
Rituraj Singh, Krishanu Saini, Anikeit Sethi, Aruna Tiwari, Sumeet Saurav, and Sanjay Singh.
\newblock Stemgan: spatio-temporal generative adversarial network for video anomaly detection.
\newblock \emph{Applied Intelligence}, 53\penalty0 (23):\penalty0 28133--28152, 2023{\natexlab{a}}.

\bibitem[Cao et~al.(2024)Cao, Lu, and Zhang]{cao2024context}
Congqi Cao, Yue Lu, and Yanning Zhang.
\newblock Context recovery and knowledge retrieval: A novel two-stream framework for video anomaly detection.
\newblock \emph{IEEE Transactions on Image Processing}, 2024.

\bibitem[Park et~al.(2022)Park, Cho, Lee, and Lee]{park2022fastano}
Chaewon Park, MyeongAh Cho, Minhyeok Lee, and Sangyoun Lee.
\newblock Fastano: Fast anomaly detection via spatio-temporal patch transformation.
\newblock In \emph{Proceedings of the IEEE/CVF Winter Conference on Applications of Computer Vision}, pages 2249--2259, 2022.

\bibitem[Ristea et~al.(2024)Ristea, Croitoru, Ionescu, Popescu, Khan, Shah, et~al.]{ristea2024self}
Nicolae-C Ristea, Florinel-Alin Croitoru, Radu~Tudor Ionescu, Marius Popescu, Fahad~Shahbaz Khan, Mubarak Shah, et~al.
\newblock Self-distilled masked auto-encoders are efficient video anomaly detectors.
\newblock In \emph{Proceedings of the IEEE/CVF Conference on Computer Vision and Pattern Recognition}, pages 15984--15995, 2024.

\bibitem[Liu et~al.(2023)Liu, Chang, Ma, Shan, and Chen]{liu2023diversity}
Wenrui Liu, Hong Chang, Bingpeng Ma, Shiguang Shan, and Xilin Chen.
\newblock Diversity-measurable anomaly detection.
\newblock In \emph{Proceedings of the IEEE/CVF conference on computer vision and pattern recognition}, pages 12147--12156, 2023.

\bibitem[Sun and Genton(2011)]{sun2011functional}
Ying Sun and Marc~G Genton.
\newblock Functional boxplots.
\newblock \emph{Journal of Computational and Graphical Statistics}, 20\penalty0 (2):\penalty0 316--334, 2011.

\bibitem[Narisetty and Nair(2016)]{narisetty2016extremal}
Naveen~N Narisetty and Vijayan~N Nair.
\newblock Extremal depth for functional data and applications.
\newblock \emph{Journal of the American Statistical Association}, 111\penalty0 (516):\penalty0 1705--1714, 2016.

\bibitem[Arribas-Gil and Romo(2014)]{arribas2014shape}
Ana Arribas-Gil and Juan Romo.
\newblock Shape outlier detection and visualization for functional data: the outliergram.
\newblock \emph{Biostatistics}, 15\penalty0 (4):\penalty0 603--619, 2014.

\bibitem[Rousseeuw and Driessen(1999)]{rousseeuw1999fast}
Peter~J Rousseeuw and Katrien~Van Driessen.
\newblock A fast algorithm for the minimum covariance determinant estimator.
\newblock \emph{Technometrics}, 41\penalty0 (3):\penalty0 212--223, 1999.

\bibitem[Hardin and Rocke(2005)]{hardin2005distribution}
Johanna Hardin and David~M Rocke.
\newblock The distribution of robust distances.
\newblock \emph{Journal of Computational and Graphical Statistics}, 14\penalty0 (4):\penalty0 928--946, 2005.

\bibitem[Bengio et~al.(2013)Bengio, Courville, and Vincent]{bengio2013representation}
Yoshua Bengio, Aaron Courville, and Pascal Vincent.
\newblock Representation learning: A review and new perspectives.
\newblock \emph{IEEE transactions on pattern analysis and machine intelligence}, 35\penalty0 (8):\penalty0 1798--1828, 2013.

\bibitem[Hinton and Salakhutdinov(2006)]{hinton2006reducing}
Geoffrey~E Hinton and Ruslan~R Salakhutdinov.
\newblock Reducing the dimensionality of data with neural networks.
\newblock \emph{science}, 313\penalty0 (5786):\penalty0 504--507, 2006.

\bibitem[Sabokrou et~al.(2016)Sabokrou, Fathy, and Hoseini]{sabokrou2016video}
Mohammad Sabokrou, Mahmood Fathy, and Mojtaba Hoseini.
\newblock Video anomaly detection and localisation based on the sparsity and reconstruction error of auto-encoder.
\newblock \emph{Electronics Letters}, 52\penalty0 (13):\penalty0 1122--1124, 2016.

\bibitem[Hasan et~al.(2016)Hasan, Choi, Neumann, Roy-Chowdhury, and Davis]{hasan2016learning}
Mahmudul Hasan, Jonghyun Choi, Jan Neumann, Amit~K Roy-Chowdhury, and Larry~S Davis.
\newblock Learning temporal regularity in video sequences.
\newblock In \emph{Proceedings of the IEEE conference on computer vision and pattern recognition}, pages 733--742, 2016.

\bibitem[Mahadevan et~al.(2010)Mahadevan, Li, Bhalodia, and Vasconcelos]{mahadevan2010anomaly}
Vijay Mahadevan, Weixin Li, Viral Bhalodia, and Nuno Vasconcelos.
\newblock Anomaly detection in crowded scenes.
\newblock In \emph{2010 IEEE Computer Society Conference on Computer Vision and Pattern Recognition}, pages 1975--1981. IEEE, 2010.

\bibitem[Huang and Sun(2019)]{huang2019decomposition}
Huang Huang and Ying Sun.
\newblock A decomposition of total variation depth for understanding functional outliers.
\newblock \emph{Technometrics}, 2019.

\bibitem[Zhang et~al.(2016)Zhang, Lu, Zhang, and Ruan]{zhang2016combining}
Ying Zhang, Huchuan Lu, Lihe Zhang, and Xiang Ruan.
\newblock Combining motion and appearance cues for anomaly detection.
\newblock \emph{Pattern Recognition}, 51:\penalty0 443--452, 2016.

\bibitem[Xu et~al.(2015)Xu, Ricci, Yan, Song, and Sebe]{xu2015learning}
Dan Xu, Elisa Ricci, Yan Yan, Jingkuan Song, and Nicu Sebe.
\newblock Learning deep representations of appearance and motion for anomalous event detection.
\newblock \emph{arXiv preprint arXiv:1510.01553}, 2015.

\bibitem[Fan et~al.(2020)Fan, Wen, Li, Qiu, Levine, and Xiao]{fan2020video}
Yaxiang Fan, Gongjian Wen, Deren Li, Shaohua Qiu, Martin~D Levine, and Fei Xiao.
\newblock Video anomaly detection and localization via gaussian mixture fully convolutional variational autoencoder.
\newblock \emph{Computer Vision and Image Understanding}, 195:\penalty0 102920, 2020.

\bibitem[Luo et~al.(2017{\natexlab{a}})Luo, Liu, and Gao]{luo2017revisit}
Weixin Luo, Wen Liu, and Shenghua Gao.
\newblock A revisit of sparse coding based anomaly detection in stacked rnn framework.
\newblock In \emph{Proceedings of the IEEE international conference on computer vision}, pages 341--349, 2017{\natexlab{a}}.

\bibitem[Li et~al.(2020)Li, Chang, and Liu]{li2020spatial}
Nanjun Li, Faliang Chang, and Chunsheng Liu.
\newblock Spatial-temporal cascade autoencoder for video anomaly detection in crowded scenes.
\newblock \emph{IEEE Transactions on Multimedia}, 23:\penalty0 203--215, 2020.

\bibitem[Wang et~al.(2018)Wang, Zeng, Liu, Zhu, Zhu, and Yin]{wang2018detecting}
Siqi Wang, Yijie Zeng, Qiang Liu, Chengzhang Zhu, En~Zhu, and Jianping Yin.
\newblock Detecting abnormality without knowing normality: A two-stage approach for unsupervised video abnormal event detection.
\newblock In \emph{Proceedings of the 26th ACM international conference on Multimedia}, pages 636--644, 2018.

\bibitem[Lv et~al.(2021)Lv, Chen, Cui, Xu, Li, and Yang]{lv2021learning}
Hui Lv, Chen Chen, Zhen Cui, Chunyan Xu, Yong Li, and Jian Yang.
\newblock Learning normal dynamics in videos with meta prototype network.
\newblock In \emph{Proceedings of the IEEE/CVF conference on computer vision and pattern recognition}, pages 15425--15434, 2021.

\bibitem[Luo et~al.(2017{\natexlab{b}})Luo, Liu, and Gao]{luo2017remembering}
Weixin Luo, Wen Liu, and Shenghua Gao.
\newblock Remembering history with convolutional lstm for anomaly detection.
\newblock In \emph{2017 IEEE International conference on multimedia and expo (ICME)}, pages 439--444. IEEE, 2017{\natexlab{b}}.

\bibitem[Hu et~al.(2022{\natexlab{b}})Hu, Yu, Wang, Zhu, Cai, and Zhu]{hu2022detecting}
Jingtao Hu, Guang Yu, Siqi Wang, En~Zhu, Zhiping Cai, and Xinzhong Zhu.
\newblock Detecting anomalous events from unlabeled videos via temporal masked auto-encoding.
\newblock In \emph{2022 IEEE International Conference on Multimedia and Expo (ICME)}, pages 1--6. IEEE, 2022{\natexlab{b}}.

\bibitem[Ning et~al.(2024)Ning, Wang, Liu, Liu, and Song]{ning2024memory}
Zhiyuan Ning, Zile Wang, Yang Liu, Jing Liu, and Liang Song.
\newblock Memory-enhanced appearance-motion consistency framework for video anomaly detection.
\newblock \emph{Computer Communications}, 216:\penalty0 159--167, 2024.

\bibitem[Zhong et~al.(2022)Zhong, Chen, Jiang, and Ren]{zhong2022cascade}
Yuanhong Zhong, Xia Chen, Jinyang Jiang, and Fan Ren.
\newblock A cascade reconstruction model with generalization ability evaluation for anomaly detection in videos.
\newblock \emph{Pattern Recognition}, 122:\penalty0 108336, 2022.

\bibitem[Tudor~Ionescu et~al.(2017)Tudor~Ionescu, Smeureanu, Alexe, and Popescu]{tudor2017unmasking}
Radu Tudor~Ionescu, Sorina Smeureanu, Bogdan Alexe, and Marius Popescu.
\newblock Unmasking the abnormal events in video.
\newblock In \emph{Proceedings of the IEEE international conference on computer vision}, pages 2895--2903, 2017.

\bibitem[Singh et~al.(2023{\natexlab{b}})Singh, Jones, and Learned-Miller]{singh2023eval}
Ashish Singh, Michael~J Jones, and Erik~G Learned-Miller.
\newblock Eval: Explainable video anomaly localization.
\newblock In \emph{Proceedings of the IEEE/CVF Conference on Computer Vision and Pattern Recognition}, pages 18717--18726, 2023{\natexlab{b}}.

\bibitem[Fang et~al.(2020)Fang, Zhou, Xiao, Li, and Yang]{fang2020multi}
Zhiwen Fang, Joey~Tianyi Zhou, Yang Xiao, Yanan Li, and Feng Yang.
\newblock Multi-encoder towards effective anomaly detection in videos.
\newblock \emph{IEEE Transactions on Multimedia}, 23:\penalty0 4106--4116, 2020.

\bibitem[Doshi and Yilmaz(2020)]{doshi2020any}
Keval Doshi and Yasin Yilmaz.
\newblock Any-shot sequential anomaly detection in surveillance videos.
\newblock In \emph{Proceedings of the IEEE/CVF conference on computer vision and pattern recognition workshops}, pages 934--935, 2020.

\bibitem[Cai et~al.(2021)Cai, Zhang, Liu, Gao, and Hao]{cai2021appearance}
Ruichu Cai, Hao Zhang, Wen Liu, Shenghua Gao, and Zhifeng Hao.
\newblock Appearance-motion memory consistency network for video anomaly detection.
\newblock In \emph{Proceedings of the AAAI conference on artificial intelligence}, volume~35, pages 938--946, 2021.

\bibitem[Li et~al.(2022)Li, Nie, Li, Zhang, and Yin]{li2022context}
Daoheng Li, Xiushan Nie, Xiaofeng Li, Yu~Zhang, and Yilong Yin.
\newblock Context-related video anomaly detection via generative adversarial network.
\newblock \emph{Pattern Recognition Letters}, 156:\penalty0 183--189, 2022.

\bibitem[Astrid et~al.(2021)Astrid, Zaheer, Lee, and Lee]{astrid2021learning}
Marcella Astrid, Muhammad~Zaigham Zaheer, Jae-Yeong Lee, and Seung-Ik Lee.
\newblock Learning not to reconstruct anomalies.
\newblock \emph{arXiv preprint arXiv:2110.09742}, 2021.

\bibitem[Ramachandra et~al.(2020)Ramachandra, Jones, and Vatsavai]{ramachandra2020learning}
Bharathkumar Ramachandra, Michael Jones, and Ranga Vatsavai.
\newblock Learning a distance function with a siamese network to localize anomalies in videos.
\newblock In \emph{Proceedings of the IEEE/CVF winter conference on applications of computer vision}, pages 2598--2607, 2020.

\bibitem[Ionescu et~al.(2019)Ionescu, Khan, Georgescu, and Shao]{ionescu2019object}
Radu~Tudor Ionescu, Fahad~Shahbaz Khan, Mariana-Iuliana Georgescu, and Ling Shao.
\newblock Object-centric auto-encoders and dummy anomalies for abnormal event detection in video.
\newblock In \emph{Proceedings of the IEEE/CVF conference on computer vision and pattern recognition}, pages 7842--7851, 2019.

\bibitem[Croitoru et~al.(2022)Croitoru, Ristea, Dascalescu, Ionescu, Khan, and Shah]{croitoru2022lightning}
Florinel-Alin Croitoru, Nicolae-Catalin Ristea, Dana Dascalescu, Radu~Tudor Ionescu, Fahad~Shahbaz Khan, and Mubarak Shah.
\newblock Lightning fast video anomaly detection via adversarial knowledge distillation.
\newblock \emph{arXiv preprint arXiv:2211.15597}, 2022.

\end{thebibliography}

\end{document}